\documentclass[runningheads]{llncs}

\usepackage{eccv}

\usepackage{eccvabbrv}

\usepackage{graphicx}
\usepackage{subcaption}

\usepackage{amsfonts}       
\usepackage{nicefrac}       
\usepackage{microtype}      

\usepackage{amsmath}
\usepackage{float}
\usepackage{placeins}

\usepackage{algorithm}
\usepackage{algpseudocode}
\usepackage{booktabs}
\usepackage{stfloats} 

\usepackage{lipsum}
\usepackage{array}  
\usepackage{multirow}
\usepackage{tabularx}
\usepackage{makecell}
\usepackage{multicol}
\usepackage{enumitem}

\usepackage[table]{xcolor}
\usepackage{bm}

\usepackage[accsupp]{axessibility}  

\usepackage{hyperref}

\usepackage{orcidlink}

\begin{document}

\title{Rare-Aware Autoencoding: Reconstructing Spatially Imbalanced Data} 

\titlerunning{Rare-Aware Autoencoding}

\author{
Alejandro Castañeda Garcia\inst{1} \and
Jan van Gemert\inst{1} \and
Daan Brinks\inst{1} \and
Nergis Tömen\inst{1}
}


\authorrunning{A.~Garcia et al.}


\institute{
Delft University of Technology (TU Delft), Delft, The Netherlands\\
\email{m.a.castanedagarcia@tudelft.nl}\\
}

\maketitle

\begin{abstract}

Autoencoders can be challenged by spatially non-uniform sampling of image content. This is common in medical imaging, biology, and physics, where informative patterns occur rarely at specific image coordinates, as background dominates these locations in most samples, biasing reconstructions toward the majority appearance. In practice, autoencoders are biased toward dominant patterns resulting in the loss of fine-grained detail and causing blurred reconstructions for rare spatial inputs especially under spatial data imbalance. We address spatial imbalance by two complementary components: (i) self-entropy-based loss that upweights  statistically uncommon spatial locations and (ii) Sample Propagation, a replay mechanism that selectively re-exposes the model to hard to reconstruct samples across batches during training. We benchmark existing data balancing strategies, originally developed for supervised classification, in the unsupervised reconstruction setting. Drawing on the limitations of these approaches, our method specifically targets spatial imbalance by encouraging models to focus on statistically rare locations, improving reconstruction consistency compared to existing baselines. We validate in a simulated dataset with controlled spatial imbalance conditions, and in three, uncontrolled, diverse real-world datasets spanning physical, biological, and astronomical domains. Our approach outperforms baselines on various reconstruction metrics, particularly under spatial imbalance distributions. These results highlight the importance of data representation in a batch and emphasize rare samples in unsupervised image reconstruction. We will make all code and related data available.


\end{abstract}

\section{Introduction}

\begin{figure}[!h]
    \centering
    \includegraphics[width=0.9\linewidth]{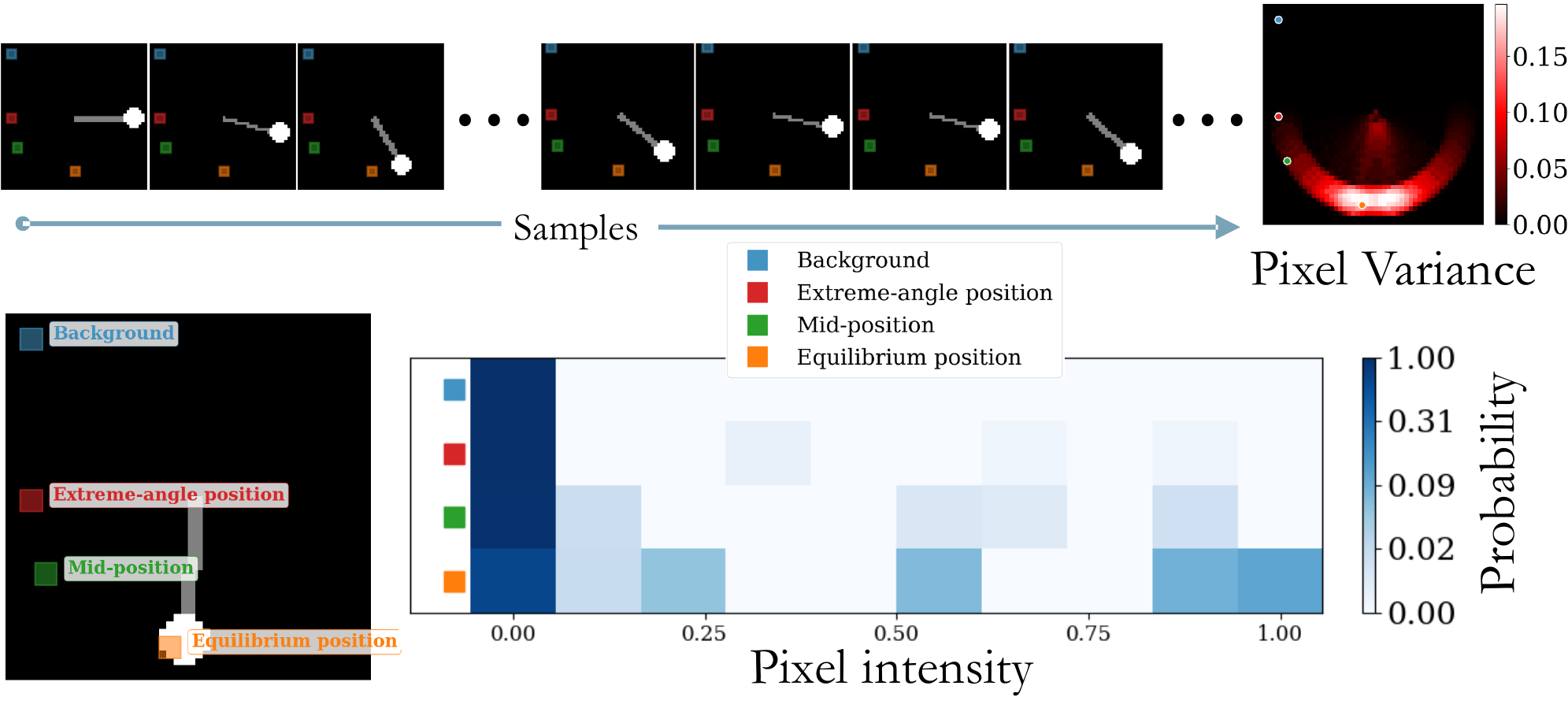}
    \caption{\textbf{Spatial data imbalance.} \textit{Top:} example frames from a pendulum dataset at different angles. Due to damping, the pendulum spends more time near equilibrium, making some spatial configurations much more frequent than others. \textit{Top-right:} per-pixel variance over the full dataset, highlighting regions with high motion-induced variability. \textit{Bottom-left:} temporal intensity traces for four representative pixels: a static background pixel (blue), an extreme-angle pixel (red), a mid-position pixel (green), and an equilibrium pixel (orange). Variance increases as pixels approach the equilibrium region, since the pendulum passes there most often. \textit{Bottom-right:} histogram of pixel intensities (for the selected pixels), showing the highly non-uniform occupancy of the state space and the resulting spatial imbalance.
    }
    \label{fig:fig0}
    \vspace{-0.6cm}
\end{figure}

We investigate small-data imaging problems in the natural sciences, where obtaining and labeling data is expensive, and given this cost, self-supervised and unsupervised learning are particularly relevant. Consequently, autoencoders are a good alternative since they do not need additional labels, as their objective is to reconstruct the input image. Additionaly, autoencoders are essential deep learning models due to their versatility across a wide range of tasks~\cite{vincent2008extracting, bengio2013representation} including denoising~\cite{vincent2008extracting}, anomaly detection~\cite{an2015variational}, latent dynamics modeling~\cite{watter2015embed}, generative modeling~\cite{diederik_p__kingma_baac31fa}, and clustering~\cite{makhzani2015adversarial} which makes them relevant for unlabeled small-data imaging problems.

Here, we focus on the autoencoder family of models for problems where the spatial locations of objects in an image varies strongly in the dataset. This is especially relevant in the natural sciences, where, \eg, stars in different galaxies have no canonical orientation and the spatial image location of cells in one petri dish, differs from the spatial image locations of cells in a different petri dish. Reasons for variance in the spatial location can be traced back to acquisition difficulties~\cite{chen_sun_f5900b3a, linus_ericsson_30c98a67, peng_zheng_d3e44641}, either for a single source~\cite{finn2017model}, or a continuous stream such as video~\cite{srivastava2015unsupervised, xiaofeng_zhang_aa4031cb}.

We make the observation that high variance, in recorded object placements over different spatial image locations in the dataset, leads to an imbalance in the unsupervised learning problem.  This imbalance is particularly present when the number of training samples is small, and objects are not equally sampled over spatial image locations in the training dataset; see Fig.~\ref{fig:fig0} for an illustration. Imbalance is a known problem~\cite{yifan_zhang_caa37b4d, charika_de_alvis_6e7a21a0, lu_yang_c3b53d32, pintea2023step}  for supervised learning when there is an imbalance in the class labels. However, in the unsupervised setting, where labels are not available, the effect of imbalance is less explored. Here, we investigate how imbalance in the spatial image locations affects autoencoders. 

In this paper we view unsupervised image reconstruction problems, for the first time, through the lens of data imbalance at spatial image locations. Inspired by methods for data imbalance in supervised settings, we design two new independent methods to handle imbalance in the unsupervised setting. One method is inspired by changing the loss and giving a higher weight to rare classes~\cite{cao2019learning,lin2017focal}. Since we don't have class labels, we introduce a self-entropy based loss to prioritize rare data. 
Similarly, another typical approach in supervised learning is emphasizing data samples~\cite{Schaul2016PER,Wu2016OnlineBootstrapping,Gu2020HardPixelMining}. Likewise, we studied sample weighting in an unsupervised setting and propose sample propagation to increase the exposure of rare inputs during training. We list here our main contributions:

\begin{itemize}   
    \item We investigate autoencoder reconstruction problems through the lens of spatial location data imbalance.
    \item We propose a self-entropy based reconstruction loss that prioritizes rare spatial pixels to avoid convergence to mean values and decrease blurring.
    \item We introduce Sample Propagation (spp), a hard-example replay mechanism, increasing the exposure of images with spatial imbalance.
    \item We, for the first time, benchmark various data balancing strategies and loss functions for autoencoders. 
\end{itemize}

\section{Related work}
\label{sec:RW}

\noindent\textbf{Image reconstruction in representation learning.} Autoencoders depend on the latent space to capture relevant image structures~\cite{srivastava2015unsupervised,hsieh2018learning, jaques2019physics, adiban2025s}. But when data distributions are biased or imbalanced, traditional loss functions often compress too aggressively or average out rare patterns. This weakens the learned representation and leads to reconstructions that are blurry or uninformative~\cite{srivastava2015unsupervised, buda2018systematic, geirhos2020shortcut}. Nevertheless, the spatial imbalance is never discussed; in contrast, we explore the effect of spatial imbalance to improve sample reconstruction to avoid blurring or averaged reconstructions.

\noindent\textbf{Reconstruction challenges due to data statistics.}
In domains like natural sciences, data from a single video often exhibits a spatial capture bias and repetitive patterns~\cite{jaques2019physics, oakden2020exploring, puyol2021fairness, garcia2024learning}. This can lead models to prioritize frequent patterns while overlooking transient or spatially rare events, resulting in blur or averaging in regions with high pixel variance~\cite{guo2023event, ercan2024hypere2vid, zhang2025reconstruction}. This imbalance leads to two main challenges. First is the averaged reconstructions, where the AE averages out or blurs details in locations with rare pixel values. Second is unbalanced samples, where the images with underrepresented rare locations are not often seen during training.

\noindent\textbf{Image blurring.}
On image reconstruction, focused losses propose plug-in alternatives which do not require architectural changes. These include pixel-level (L1, L2), perceptual (VGG~\cite{zhang2018unreasonable}, Watson~\cite{czolbe2020watson}), frequency-aware (FFL~\cite{jiang2021focal}) and class-balanced (LDAM~\cite{cao2019learning}, Focal~\cite{lin2017focal}) losses. Similar, Bredell et al.~\cite{bredell2023explicitly} introduced a blur-aware loss (BEL) for variational autoencoders (VAEs) targeting high-frequency errors while preserving ELBO maximization. While these methods offer diverse reconstruction strategies, they have not been compared and evaluated under sample imbalance. Here, we provide such an evaluation in comparison to our proposed method, to properly formalize the spatial imbalance problem in AE reconstruction and show whether the methods dealing with blurriness aid the underlying problem.

\noindent\textbf{Imbalanced sample distribution in supervised settings.} Data imbalance is extensively studied in classification~\cite{yifan_zhang_caa37b4d, pintea2023step}, where the goal is to ensure minority classes are properly represented. However, data imbalance has received less attention in reconstruction. Some methods like Online Hard Example Mining~\cite{Shrivastava2016OHEM}, Prioritized Experience Replay~\cite{Schaul2016PER}, and hard-pixel mining~\cite{Wu2016OnlineBootstrapping,Gu2020HardPixelMining} are  ill-suited in a unsupervised setup since their notion of “hardness” depends on previous supervised annotations. Other methods can be transferred to unsupervised settings such as, loss function modifications, like reweighting sample contributions~\cite{cao2019learning,lin2017focal}, have proven effectiveness. Likewise, Loss-based solutions such as focal loss~\cite{lin2017focal}), or margin-based adjustments such as ldam~\cite{cao2019learning}. Beyond loss adjustments, strategies such as online batch selection~\cite{loshchilov2015online}, curriculum learning~\cite{bengio2009curriculum} and selection via proxy~\cite{Coleman2020SVP} dynamically prioritize harder samples. Mentioned approaches suggest that imbalance can be addressed through both loss design and data sampling. However, their application to autoencoders remains largely unexplored--particularly within the batch learning paradigm~\cite{goodfellow2016deep}, which is central to modern deep learning pipelines. Ignoring batch-level dynamics can bias gradient estimates, underrepresented rare patterns, and degrade generalization in reconstruction. While the existing methods work on their particular task, we studied them together, for the first time, in a unsupervised setup for image reconstruction with spatial imbalanced data.

\section{Methods}


\begin{figure}[!h]
    \centering
    \includegraphics[width=0.7\linewidth]{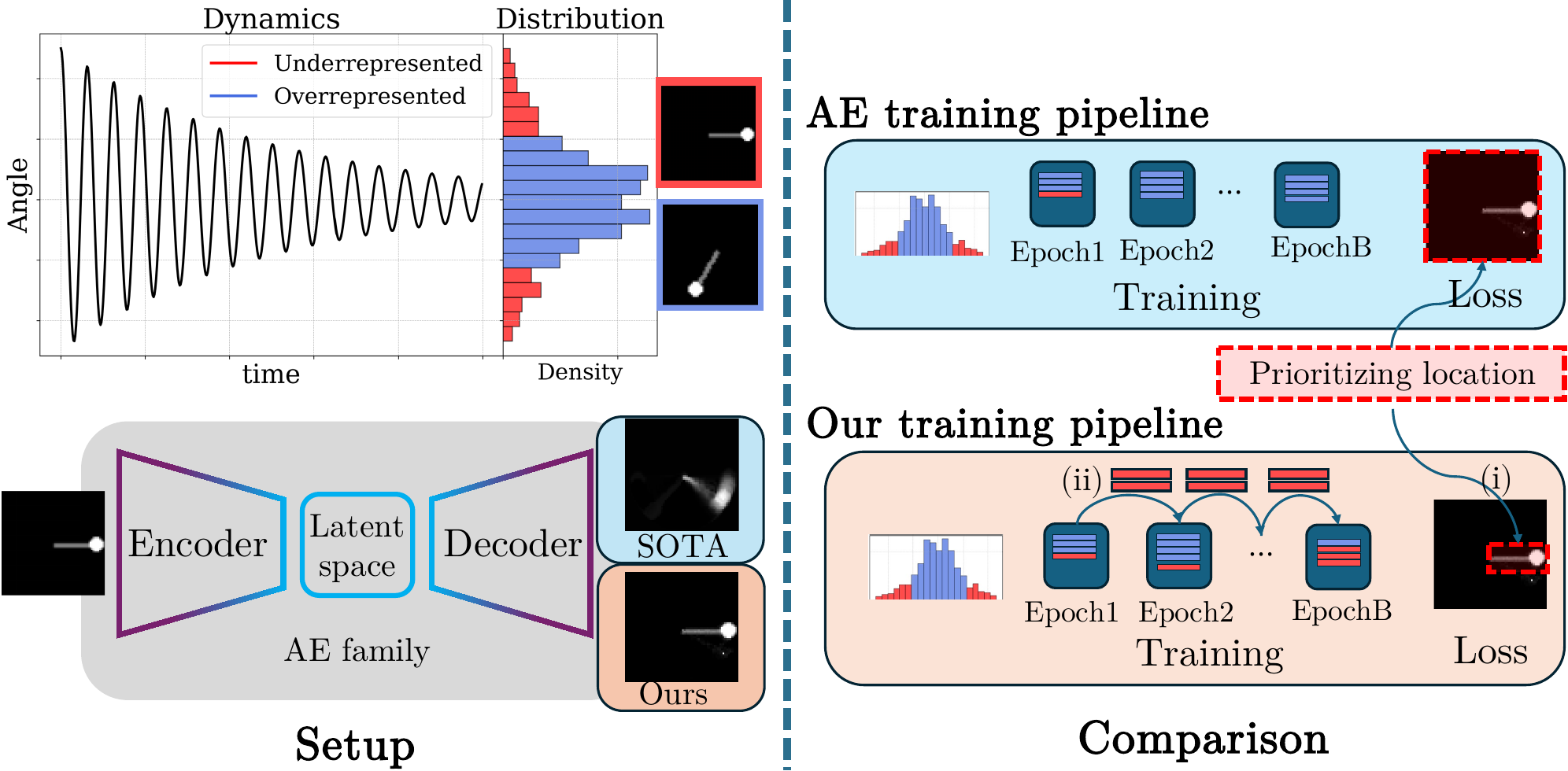}
    \caption{
        \textbf{Addressing reconstruction under spatial data imbalance.} Consider a dataset of simulated pendulum dynamics. Black squares are input or reconstructed frames. \textbf{Left-top:} Input distributions exhibit a strong bias, with common, overrepresented samples (blue) and underrepresented ones (red). \textbf{Left-bottom:} Standard autoencoder (AE) backbone (encoder–latent–decoder). The SOTA (cyan)  approach tends to converge toward smooth reconstructions dominated by frequent samples. \textbf{Right:} Our approach (orange) introduces two complementary mechanisms:
        (i) a \emph{self-entropy loss} that upweights high-surprisal pixels across the batch within each image, and
        (ii) \emph{sample propagation} that replays high-loss (hard) samples across epochs, ensuring rare
        events are repeatedly observed. 
         The combined effect yields sharper and more balanced reconstructions.}
    \label{fig:fig1}
    \vspace{-0.6cm}
\end{figure}

\paragraph{Self-entropy loss}
Our loss aims to decrease the blurriness caused by converging to the average pixel values. The intuition is to assign greater importance to pixel values that are statistically rare across a batch, in theory we are interested in the full distribution, but in practice we look at the batch level since we do not want to weight spatial locations that the model won't see in a particular update. The model is encouraged to prioritize spatially imbalanced locations over statistically redundant locations. Specifically, we first measure the self-entropy, associated with each pixel across the batch, those pixel forms the rare location to prioritize denominated as "entropy mask".

For image batch $X$ and its reconstruction $\hat X$, with \(X,\hat X \in [0,1]^{B\times C\times H\times W}\), batch size $B$, number of channels $C$, width $W$ and height $H$, we flatten spatial and channel dimensions so \(x_{b,\ell}\) indices sample \(b\in\{1,\dots,B\}\) and location \(\ell\in\{1,\dots,L\}\) with \(L=CHW\). We estimate the distribution using a histogram with \(J\in\mathbb{N}\) bins and bin width \(\delta = 1/J\). We discretize each \(x_{b,\ell}\) to a bin $k_{b,\ell} = \big\lfloor x_{b,\ell} \cdot (J-1)\big\rfloor \in \{0,\dots,J-1\}$ and count occurrences $c_{\ell,k}$ of pixel values in the batch,
$c_{\ell,k} = \sum_{b=1}^{B} \mathbf{1}[\,k_{b,\ell}==k\,]$ to estimate the pixel value probability (more details in Appendix~\ref{sec:apdx_pdf})
\vspace{-0.2\baselineskip}
\begin{equation}
\widehat{p}_\ell(x_{b,\ell}) \approx \frac{c_{\ell,k}/B}{\delta}
= \frac{J}{B}\,c_{\ell,k}.\vspace{-0.5\baselineskip}
\end{equation}
We calculate the self-entropy of each pixel $s_{b,\ell}$ 
\vspace{-0.3\baselineskip}
\begin{equation}
s_{b,\ell} = -\log\!\big(\widehat{p}_\ell(x_{b,\ell}) + \varepsilon\big),\vspace{-0.3\baselineskip}
\end{equation}
using \(\varepsilon>0\) for numerical stability.
We reshape back to \((B,C,H,W)\),  and apply per-(image,channel) min–max normalization, and define an "entropy mask" as $\tilde s$:
\vspace{-0.2\baselineskip}
\begin{subequations}
\begin{align}
m_{b,c} \!=\! \min_{h,w} s_{b,c,h,w},\quad
M_{b,c} \!=\! \max_{h,w} s_{b,c,h,w}
\;\Rightarrow\;\\
\tilde s_{b,c,h,w}
= \frac{s_{b,c,h,w} - m_{b,c}}{M_{b,c} - m_{b,c} + \varepsilon} \in [0,1].\vspace{-0.5\baselineskip}
\end{align}
\end{subequations}

where $b,c,h,w$ are indices for each pixel in $[1, \dots, B]$, $[1, \dots, C]$, $[1, \dots, H]$, $[1, \dots, W]$ respectively.


We define our self-entropy loss as a spatially weighted L1 distance. Let the pixel-wise error be: \(\text{L1}_{b,c,h,w}=\lvert \hat X_{b,c,h,w}-X_{b,c,h,w}\rvert\) The corresponding self-entropy weight is defined as: ${{\text{ent}}_{b,c,h,w} = \tilde s_{b,c,h,w} + 0.05}$.  where $\text{L1} \in [0,1]$ and $\tilde s \in [0,1]$. Intuitively, the L1 term encourages the recovery of sharp details, while the self-entropy weight $\text{ent} \in [0.05, 1.05]$ identifies which pixels to prioritize. The constant 0.05 acts as a smoothing factor to ensure that common pixels are not completely ignored during training. To ensure the model reconstructs both rare features and the global image structure, we balance the standard Mean Squared Error (MSE) with our proposed self-entropy loss:

\vspace{-0.3\baselineskip}
\begin{subequations}
\begin{align}
\mathrm{Ent}_b &= \tfrac{1}{CHW}\!\sum_{c,h,w} \text{L1}_{b,c,h,w}\, {\text{ent}}_{b,c,h,w},\\
\mathrm{
\vspace{-0.2\baselineskip}MSE}_b&= \tfrac{1}{CHW}\!\sum_{c,h,w} \big(\hat X_{b,c,h,w}-X_{b,c,h,w}\big)^2,\vspace{-0.5\baselineskip}
\end{align}
\end{subequations}
and combine with a trade-off coefficient \(\lambda>0\):
\vspace{-0.3\baselineskip}
\begin{equation}
    \label{eq:entropy_loss}
        \ell_b = \mathrm{MSE}_b + \lambda\,\mathrm{Ent}_b,\quad
        \mathcal{L} = \tfrac{1}{B}\sum_{b=1}^{B} \ell_b.
\end{equation}

\paragraph{Sample propagation (Spp)}


Entropy alone is not enough to solve the problem, while self-entropy prioritizes a particular location, the model may forget it if the samples with this data imbalance are not seen again. To counter data imbalance, we propagate difficult samples across subsequent batches. We find this more favorable to reweighting individual samples per batch as traditional focal methods.

Our algorithm identifies samples with rare, difficult-to-reconstruct spatial locations and propagates them into subsequent batches to prevent the model from "forgetting" these features. To avoid overfitting to these specific patterns, we restrict the model update to the top $B$ hardest samples. This ensures that while the effective batch size processed by the model increases, the number of samples used for gradients remains constant, maintaining training efficiency.

\begin{figure}[t]
    \centering
    \includegraphics[width=0.9\linewidth]{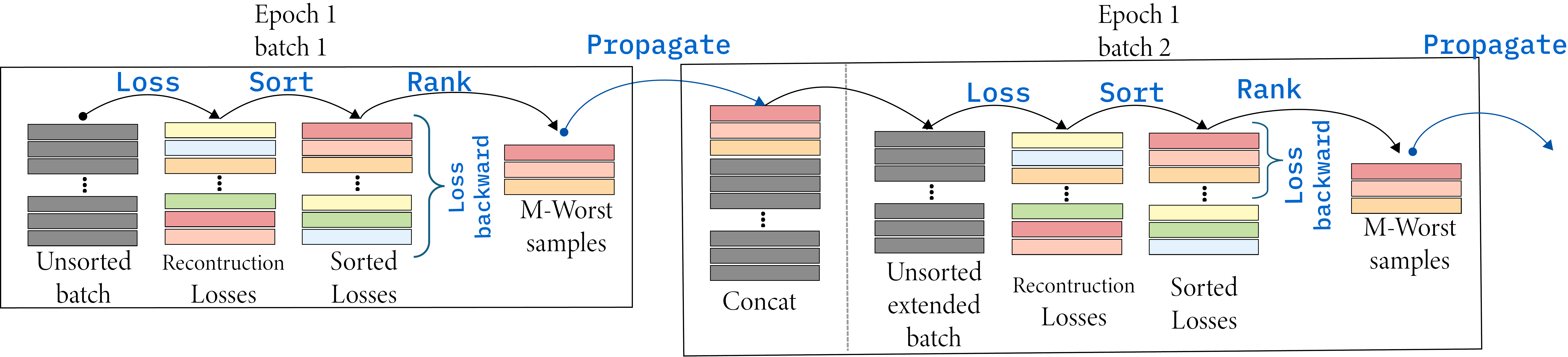}
    \caption{\textbf{Sample propagation algorithm.} First, for each input sample, the model produces a reconstruction error. The color scale encodes error magnitude: red denotes higher error, while decreasing progressively to cooler hues. Second, we compute a error per sample and sort/rank the batch accordingly, yielding an ordered list from highest to lowest reconstruction error. Finally, a propagation step selects the $M$ top-ranked (hardest) samples for targeted optimization and concatenates them to the next batch; gradients are then applied (loss backward) to update the network with an emphasis on hard examples.}
    \label{fig:Spp}
    \vspace{-0.2cm}
\end{figure}

For a particular batch $\text{b}_i$ with size $B$ during training: 
1) Reconstruct: Compute the reconstruction for all samples in batch $\text{b}_i$. 
2) Evaluate: Sort samples based on their individual reconstruction error.
3) Identify: Select the $M$ most difficult samples as "hard" cases.
4) Propagate: Concatenate these $M$ samples into the next training batch $\text{b}_{i+1}$.
5) Update: Perform a gradient update using only the $B$ hardest samples from the resulting pool.
Sample propagation is presented in Fig.~\ref{fig:Spp}.


We use the following technique inspired by the focal-loss~\cite{lin2017focal}. With \linebreak \mbox{\(\ell_{\min}=\min_b \ell_b\)}, \(\ell_{\max}=\max_b \ell_b\), define
\begin{equation}
f_b = 1+ \frac{\ell_b-\ell_{\min}}{\max(\ell_{\max}-\ell_{\min},\,\varepsilon)} (B-1) \in [1,B],\\
\end{equation}



\noindent where the final loss is \vspace{-0.5cm}
\begin{equation}
\mathcal{L} = \tfrac{1}{B}\sum_{b=1}^{B} f_b\,\ell_b.
\end{equation}

\section{Experiments}

To ensure a fair comparison, we set the same training hyperparameters across baselines and proposed methods. The models are trained on an NVIDIA A40 GPU using the Adam optimizer with a learning rate $lr = 1.0e^{-3}$ and weight decay $wd = 1.0e^{-5}$. For full architectural details, refer to the supplementary material.

We evaluate using Mean Squared Error (MSE↓), Peak Signal-to-Noise Ratio (PSNR↑), and Structural Similarity Index (SSIM↑), following prior work~\cite{mustafa2022training, bredell2023explicitly}. 


For notation, we denote our sample propagation method as spp$k$  related to memory size $M$ as follows: $M = \frac{B}{k} $ . For $\text{spp}k$ algorithm, our internal experiments showed that a fixed amount of memory was more stable than a dynamic memory size; therefore, we compare different memory sizes. For simplicity and readability we refer to "self-entropy" just as "entropy" and experiments denoted with "spp+entropy" we used $k = 4$. 

\subsection{Experiments on controlled data}

\begin{figure}[t]
    \centering
    \includegraphics[width=0.8\linewidth]{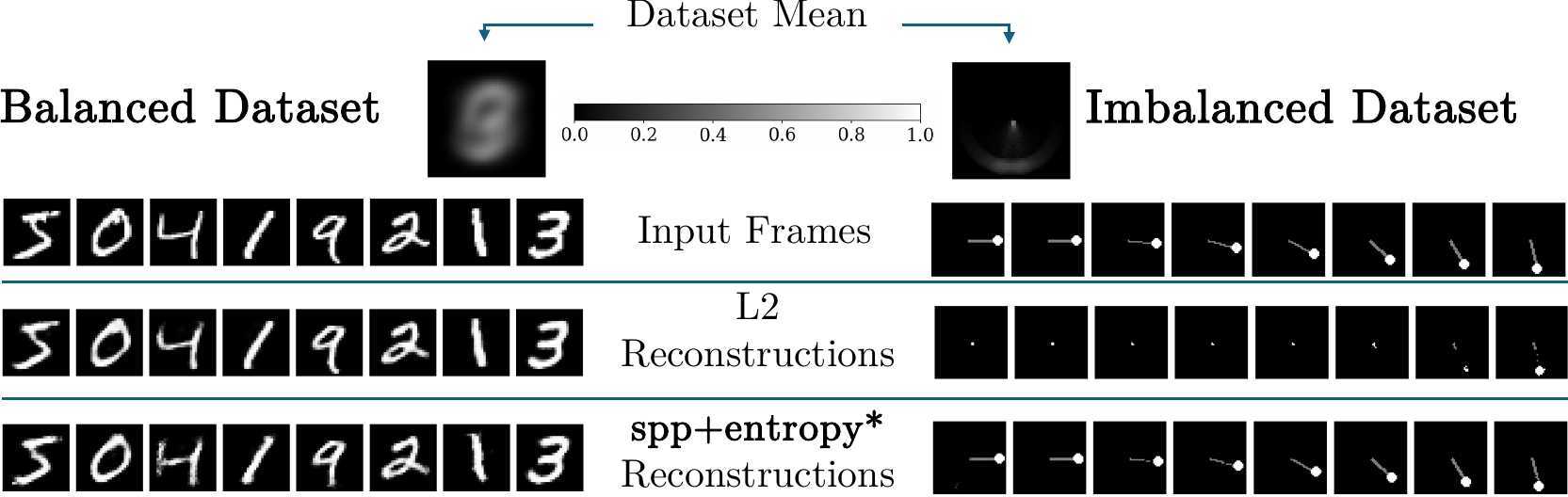}
    \caption{ \textbf{Comparison between balanced and imbalanced reconstruction scenarios.}
    The MNIST dataset (left) represents a balanced case where samples are more uniformly distributed across the spatial domain. In contrast, the damped pendulum dataset (right) introduces imbalance, as certain image spatial locations are severely underrepresented across the dataset and within training batches. Our proposed spp+entropy framework reconstructs both balanced and imbalanced data faithfully, while the standard MLP AE pipeline performs well only on homogeneous datasets such as MNIST but fails to preserve structure under imbalance.
    }
    \label{fig:ImbvsBal} 
    \vspace{-0.4cm}    
\end{figure}

We use a simulated dataset~\cite{garcia2024learning} of a damped pendulum, with a size of $(64 \times 64)$ pixels, where the black background and the damping leads to spatial imbalance and blurry reconstructions. We use MNIST as an example of a  homogeneous data distribution with uniform spatial coverage. Results for a fully connected (MLP) AE encoder and decoder are shown in Fig.~\ref{fig:ImbvsBal}. For the imbalanced pendulum, the autoencoder overfits to dominant spatial locations, leading to incomplete or distorted reconstructions. In contrast, our spp+entropy method maintains reconstruction fidelity in both cases, demonstrating robustness to spatial imbalance distributions that otherwise degrade in conventional reconstruction approaches.

\begin{figure}[t]
    \centering
    \includegraphics[width=0.97\linewidth]{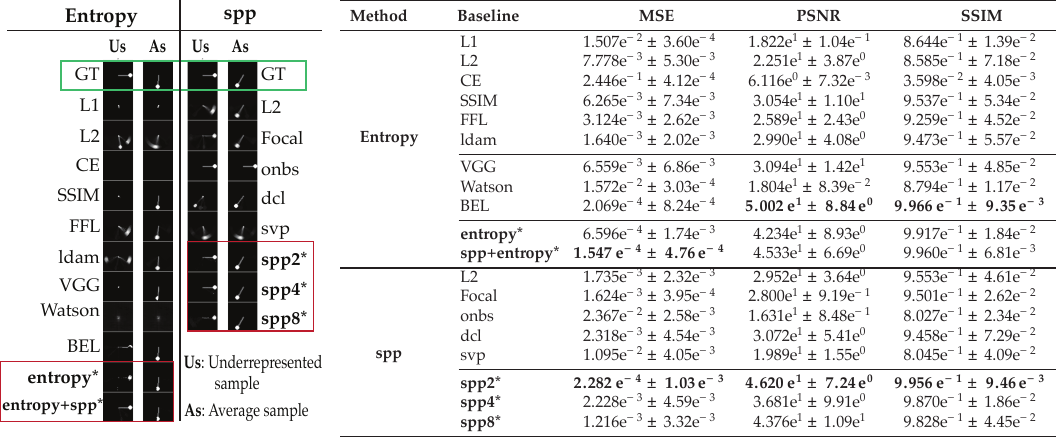}
    \captionof{figure}{
    \textbf{(\textit{Left}) Qualitative comparison.} The first row shows the ground truth (GT). Methods are arranged by rows: (Left) comparison of baselines and our entropy-based methods (\textbf{entropy*}, \textbf{entropy+spp*}); (right) baselines and our underrepresented-sample-focusing spp variants (\textbf{spp2*}, \textbf{spp4*}, \textbf{spp8*}). Columns include both an \emph{Underrepresented sample} (Us) and an \emph{Average sample} (As) to visualize behavior on rare versus frequent cases. The layout highlights how different objectives affect sharp structures and textures: the left column makes visible the tendency of some losses to approach a dataset-average appearance, whereas the \emph{spp} block illustrates how prioritizing underrepresented samples influences detail retention and stability across memory sizes $k$.\label{fig:qualToy}
    }\vspace{-0.35cm}
    \captionof{table}{\textbf{(\textit{Right}) Reconstruction performance on the controlled dataset.} The table reports reconstruction metrics (MSE↓, PSNR↑, SSIM↑; mean$\pm$std over samples) for two complementary proposals: \emph{(i) Entropy} (top block), which evaluates losses under a setting prone to convergence toward a dataset’s statistical mean, and \emph{(ii) spp} (bottom block), which benchmarks losses against our sample-propagation  with different memory sizes (spp2, spp4, spp8) which prioritizes underrepresented samples. $*$ denotes our methods; entries in \textbf{bold} indicate the best mean value within each experiment block for a given metric. Our methods demonstrate better qualitative performance at reconstruction while keeping a high statistical score across the whole dateset.}    
    \label{tb:toyResults}
    \vspace{-0.4cm}    
\end{figure}

In Fig.~\ref{fig:qualToy} and Table~\ref{tb:toyResults} we evaluate various baselines. Most baselines target fidelity and perceptual agreement: pixel losses (L1/L2), SSIM for structural similarity \cite{Wang2004SSIM}, perceptual (VGG-feature) losses \cite{zhang2018unreasonable}, the Watson DCT-based perceptual error model \cite{czolbe2020watson}, frequency-aware FFL \cite{jiang2021focal}, and margin/imbalance-aware ldam \cite{cao2019learning}. Finally the recent BEL (Blur Error Loss)~\cite{bredell2023explicitly} introduced by Bredell et al. Baseline methods for our entropy loss explicitly penalize error in pixel/structure/frequency spaces or encourage larger margins on minority patterns. For the sample-propagation baselines adapt the sampling or loss to class/instance rarity: Focal loss emphasizes hard/rare examples \cite{lin2017focal}; curriculum learning schedules data from easy to hard \cite{bengio2009curriculum}; online batch selection (onbs) prefers high-loss items within each step \cite{loshchilov2015online}; and selection via proxy (svp) uses a lightweight proxy model to guide which samples to revisit \cite{Coleman2020SVP}.

As seen in Fig.~\ref{fig:qualToy}, across datasets and architectures, our two methods provide complementary benefits: the entropy-based loss counteracts collapse toward mean appearances, recovering sharper details and higher PSNR / lower MSE, while the underrepresented-sample focusing (spp$k$) extends the exposition of rare samples. We note that the BEL baseline attains  strong scores on the toy setting aided by its sensitivity to the synthetic, high-contrast edges—yet this advantage diminishes on real data where textures and structures are more varied (see Sec.~\ref{sec:realExp}). The qualitative comparisons in Fig.~\ref{fig:qualToy} highlights the differences: our methods preserve fine structure and rare patterns that baselines tend to smooth, translating the quantitative gains into more faithful reconstructions.

\begin{figure}[t]
    \begin{subfigure}[]{0.47\textwidth}        
        \includegraphics[width=0.98\linewidth]{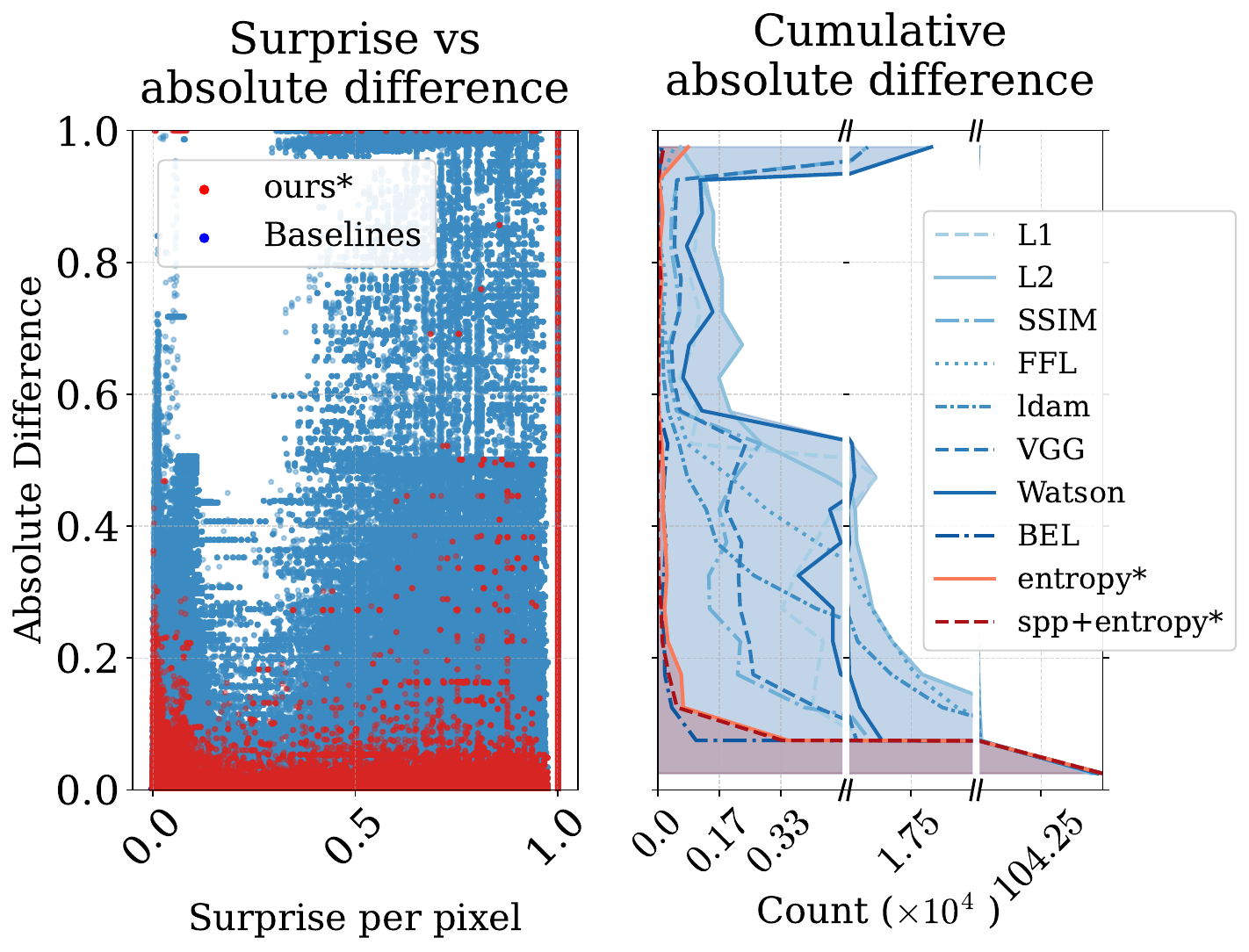}
        \caption{\textbf{Per-pixel error vs.\ surprise with marginal distributions.} (Left) Scatter–density plots show absolute reconstruction error against per-pixel surprise (higher = rarer/less expected) for baselines and our methods (\textbf{entropy*}, \textbf{spp+entropy*}). The joint distributions for our methods concentrate near the origin—low error even at moderate–high surprise—while baselines spread more uniformly across surprise values. (Right) The accompanying marginal histograms highlight that our curves accumulate most mass at small absolute differences, with markedly reduced heavy tails compared to baselines.}
        \label{fig:surpvsdiff}
        \vspace{-0.2cm}
    \end{subfigure}
    \hspace{0.02\textwidth}
    \begin{subfigure}[]{0.51\textwidth}        
        \includegraphics[width=0.98\linewidth]{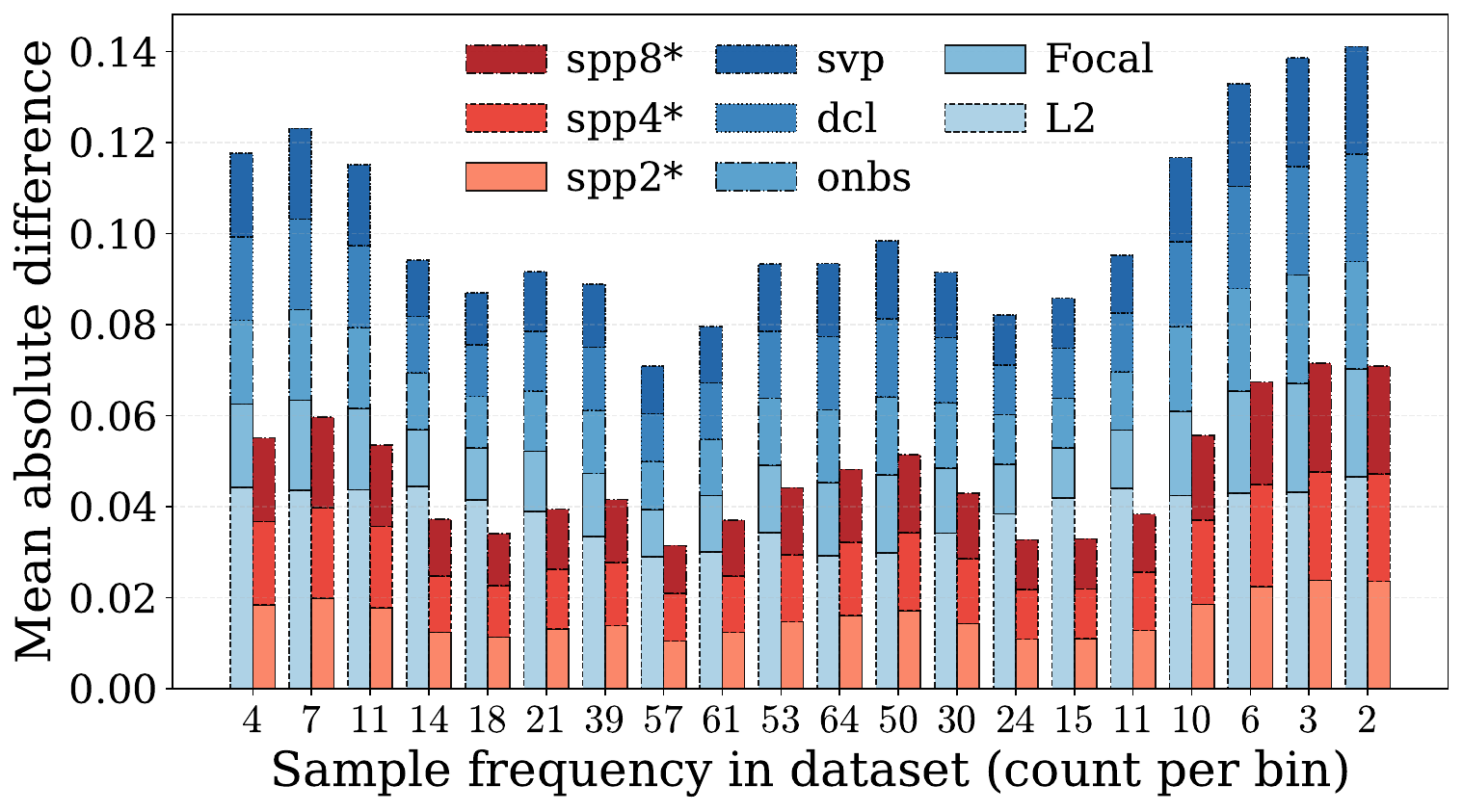}
        \caption{\textbf{Error vs.\ sample frequency (grouped by occurrence count).} Mean absolute difference against binned sample frequency in the dataset for baselines (in blue) and our sample propagation variants (in red: \textbf{spp2*}, \textbf{spp4*}, \textbf{spp8*}).
        Baselines are overlapped in a single column as well as our spp variants to improve readability.
        Our curves remain both \emph{lower} and \emph{flatter} across the frequency axis—indicating uniformly low reconstruction error—even in the rarest bins. In contrast, baseline errors rise as frequency decreases, revealing sensitivity to underrepresented cases.}
        \label{fig:labeldist}
        \vspace{-0.5cm}
    \end{subfigure}  
    \caption{Hypothesis testing on synthetic data. On figure (a) we show how our entropy loss reduce the error of individual pixel equally, while on (b) sample propagation reduces the error on underrepresented samples having a similar performance in all samples. }
\end{figure}

Furthermore, Fig.~\ref{fig:surpvsdiff} indicates that our objective reduces average error across all pixels and  maintains low error on pixels with high surprise—precisely where models typically fail. In contrast, standard losses distribute error more evenly over the surprise axis, leaving a sizable fraction of high-surprise pixels with larger residuals. 

Fig.~\ref{fig:labeldist} shows that propagating underrepresented samples yields improvements that are \emph{global} (lower average error across all counts) and \emph{targeted} (especially strong in low-frequency bins). Whereas conventional objectives exhibit a pronounced error ramp as samples become rarer, our spp$k$ variants maintain a nearly flat error profile—evidence that the model learns features that transfer to tail examples rather than overfitting to frequent modes. Among our methods, larger memories (spp2*, spp4*) further stabilize the tail without sacrificing performance on common cases, supporting the claim that controlled replay of rare patterns mitigates frequency-induced degradation and improves robustness across the full data distribution.

\begin{figure}[t]
    \centering
    \includegraphics[width=0.5\linewidth]{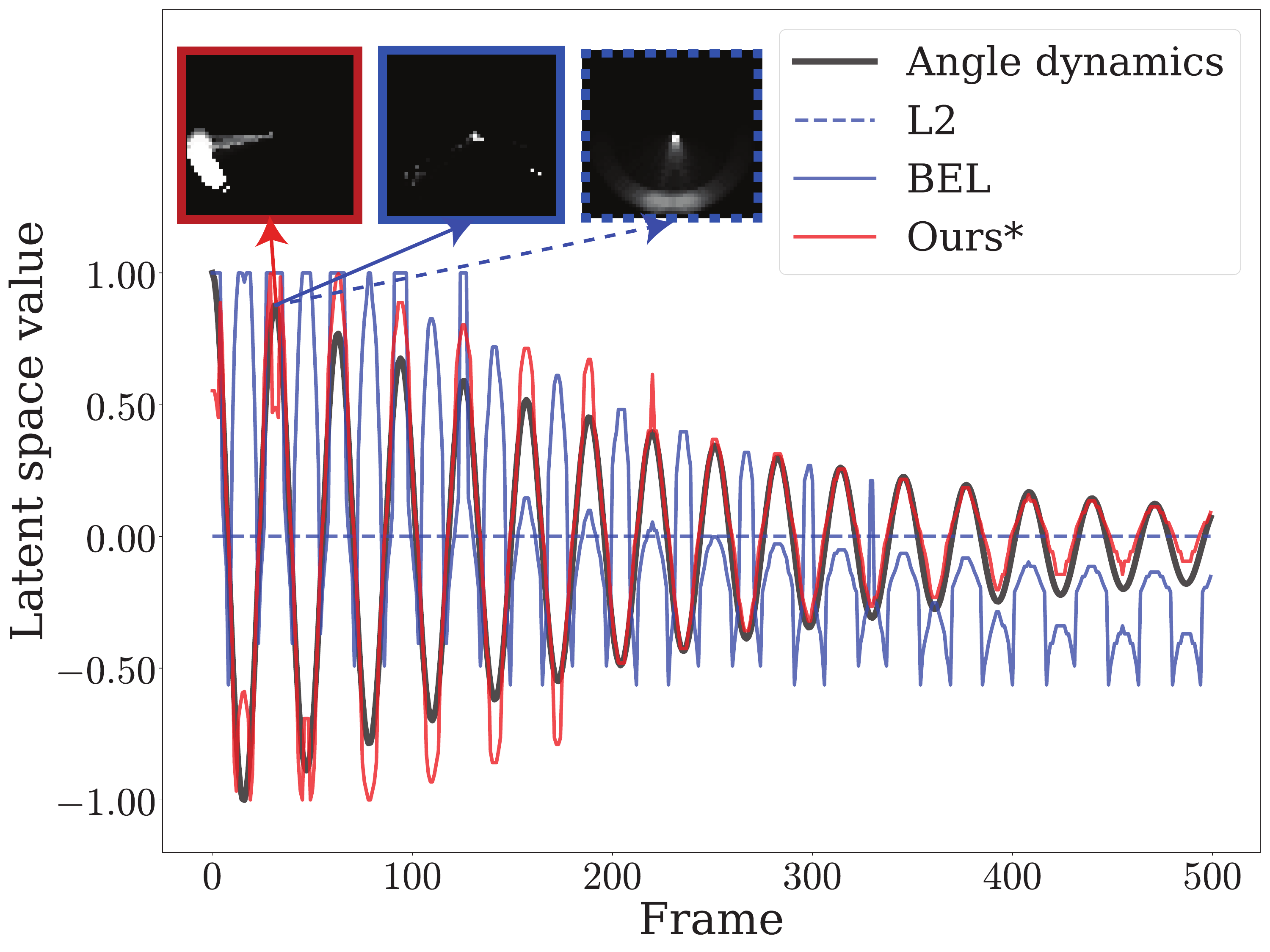}
    \caption{
    \textbf{Latent–capacity comparison under restricted bottlenecks.} Expressivity in the AE latent space when its dimensionality is intentionally constrained to 1D. We report the time–series plots that trace the decoded signal across samples. Under tight bottlenecks, baseline trajectories either collapse (L2) or do not follow the oscillations faithfully (BEL), whereas \textbf{our method} (spp+entropy) has rich latent dynamics that remain coupled to the input sequence. Example reconstructions from each method (upper left) illustrate it.
    }
    \label{fig:compLat}
    \vspace{-0.5cm}
\end{figure}

Finally, in Fig.~\ref{fig:compLat} we visualize a restricted latent space, using only one-dimensional (1D) space. Our method has lower MSE/higher PSNR with comparable or reduced variance, which  is visually apparent in the trajectories: the L2 loss leads to latent space collapse, while BEL partly captures the oscillatory dynamics, but does not yield a faithful trajectory or detailed reconstructions. In contrast, our method remains expressive of the system’s salient, time-varying structure. In short, under restricted capacity, our objective resists premature convergence and captures dynamics more aligned with the real dynamics.

\subsection{Real data experiments}
\label{sec:realExp}


We used three different datasets which have spatial distribution imbalance or imbalance due to rare events. 

\noindent\textit{Pendulum:} A real-world pendulum video~\cite{garcia2024learning}. A static background and recurring ball positions dominate the frame distribution, while some ball positions appear infrequently due to the pendulum’s damping.\\
\noindent\textit{Galaxy zoo:} A dataset of galaxy images spanning diverse morphological types~\cite{lintott2011galaxy}. The dataset features a uniformly black background and predominantly centered galaxies. Off-center galaxy images and examples that showcase fine structural details (like stars) occur infrequently.\\
\noindent\textit{Mitosis:} Time-lapse microscopy sequence for cell tracking and mitosis detection from the MOTChallenge CTMC-v1 dataset~\cite{anjum2020ctmc}.The dataset shows uniform imaging conditions and interphase cell instances, while mitotic events appear infrequently due to their brief duration.

We evaluate the following autoencoder variants: An \textit{MLP Autoencoder (AE)}~\cite{chen2023auto} consisting of fully connected architecture composed of a multilayer perceptron (MLP) encoder and decoder, 
a \textit{Variational Autoencoder (VAE)}~\cite{diederik_p__kingma_baac31fa,chen2023auto}, 
a \textit{Sparse Autoencoder (SAE)}~\cite{chen2023auto}, 
and a \textit{Masked Autoencoder (MAE)}~\cite{he2022masked}. 

\textit{Entropy:} We analyzed the performance of the proposed entropy loss function (Eq.~\ref{eq:entropy_loss}) qualitatively compared to baseline loss functions (Fig.~\ref{fig:qualConver}). Quantitatively, tables \ref{tb:PendulumResultsConvergence}, \ref{tb:GalaxyResultsConvergence}, \ref{tb:MitosisResultsConvergence} (in the appendix) are summarized in Fig.~\ref{fig:summ_conv} which shows the performance of our method.
For the quantitative evaluation, we compare standard pixel-wise loss functions  (L1, L2, CE), and structural (SSIM), frequency/feature (FFL), margin-aware (ldam) and perceptual (VGG, Watson, BEL) losses against our proposed \emph{entropy*} loss and its variant \emph{spp+entropy*} on four backbones tailored to the experiment (AE, VAE, SAE, MAE). 

\begin{figure}[H]
    \centering
    \begin{subfigure}[]{0.85\textwidth}
        \centering
        \includegraphics[width=\linewidth]{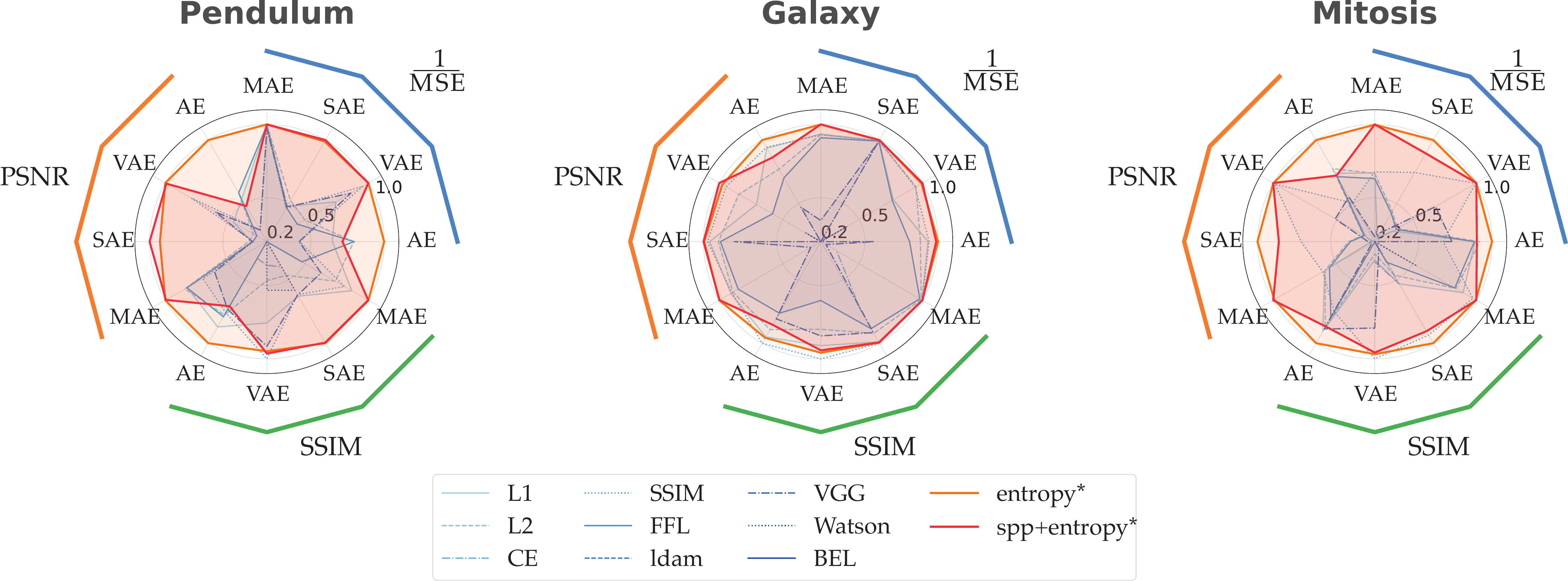}
        \caption{
        \textbf{Reconstruction results across datasets and architectures with different losses.} Across all datasets and architectures, our losses consistently achieve the best distortion metrics (lowest MSE, highest PSNR), with frequent SSIM ties or improvements. On repetitive dynamics (\textit{Pendulum}), entropy–based training prevents convergence to oversmoothed solutions, yielding MSE reductions over strong perceptual/pixel baselines. On structured textures (\textit{Galaxy}), it improves PSNR and MSE  while remaining SSIM–competitive. On fine cellular detail (\textit{Mitosis}), it preserves morphology, improving PSNR and reducing MSE in AE/SAE. Overall, maximizing information content during training generalizes across regimes, sharpening the reconstruction and avoiding collapse. \vspace{0.2cm}
        }
        \label{fig:summ_conv}
    \end{subfigure}    
    \hfill
    \begin{subfigure}[]{0.8\textwidth}
        \centering
        \includegraphics[width=\linewidth]{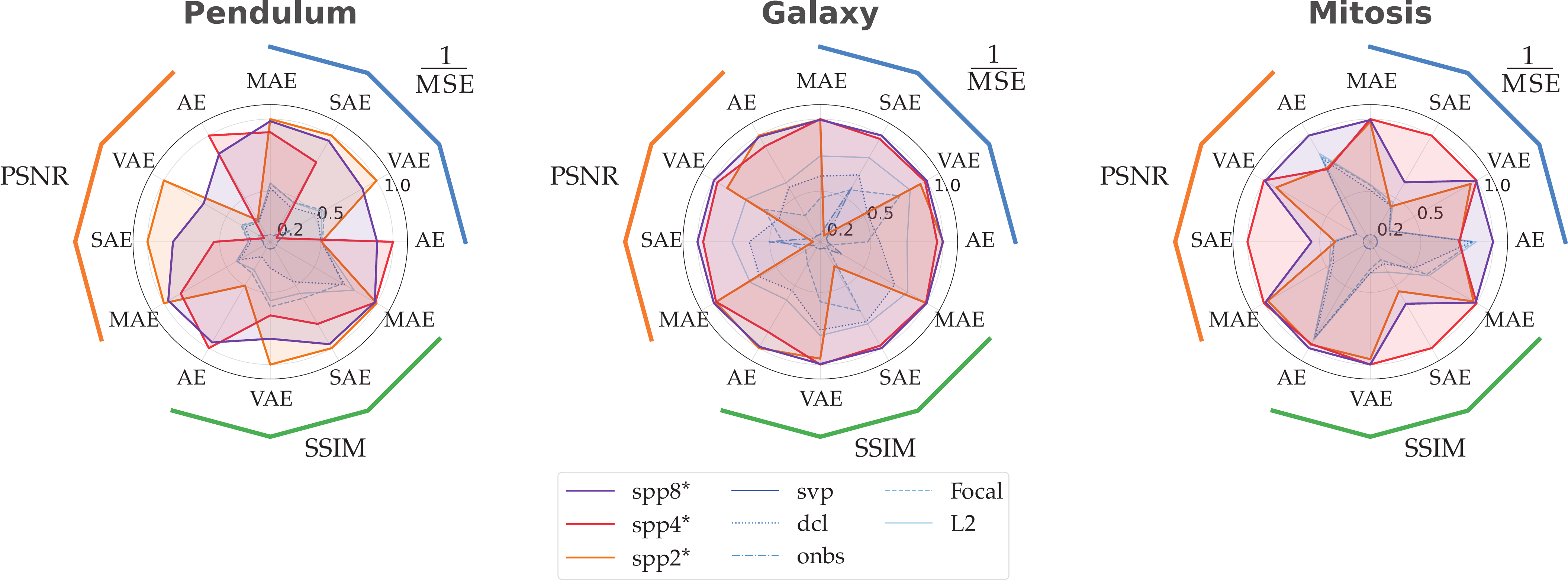}
        \caption{
        \textbf{Sample focus reconstruction performance across datasets and architectures.} Overall, $\text{spp}k$ improves average reconstruction quality relative to pixel/perceptual baselines while keeping the standard deviation comparable or lower—indicating better robustness to rare/atypical samples. Taken together, the figure shows that focusing on underrepresented samples improves the general mean–variance reconstruction in favor of our model.
        }
        \label{fig:summ_spp}
    \end{subfigure}
    \caption{
    Summary figure of our extensive results detailed in appendix tables~\ref{tb:PendulumResultsConvergence} to \ref{tb:MitosisResultsSpp}. Fig.~\ref{fig:summ_conv} shows the performance of our entropy* loss function and \ref{fig:summ_spp} the results for our \textbf{spp} algorithm. Both sub-figures show the performance on three different datasets \textit{Pendulum}, \textit{Galaxy}, and \textit{Mitosis} over four autoencoder families (AE, VAE, SAE, MAE).  For better visualization, we re-normalized the metrics from $ 0.2 \; \textrm{to} \; 1.0$ where high is always better. In the presented radar plots more area implies a better overall reconstruction. Baseline methods are shown in shades of blue while our methods are in a red scale.}
    \label{fig:main}
\end{figure}

\begin{figure}[t]
    \centering
    \includegraphics[width=0.7\linewidth]{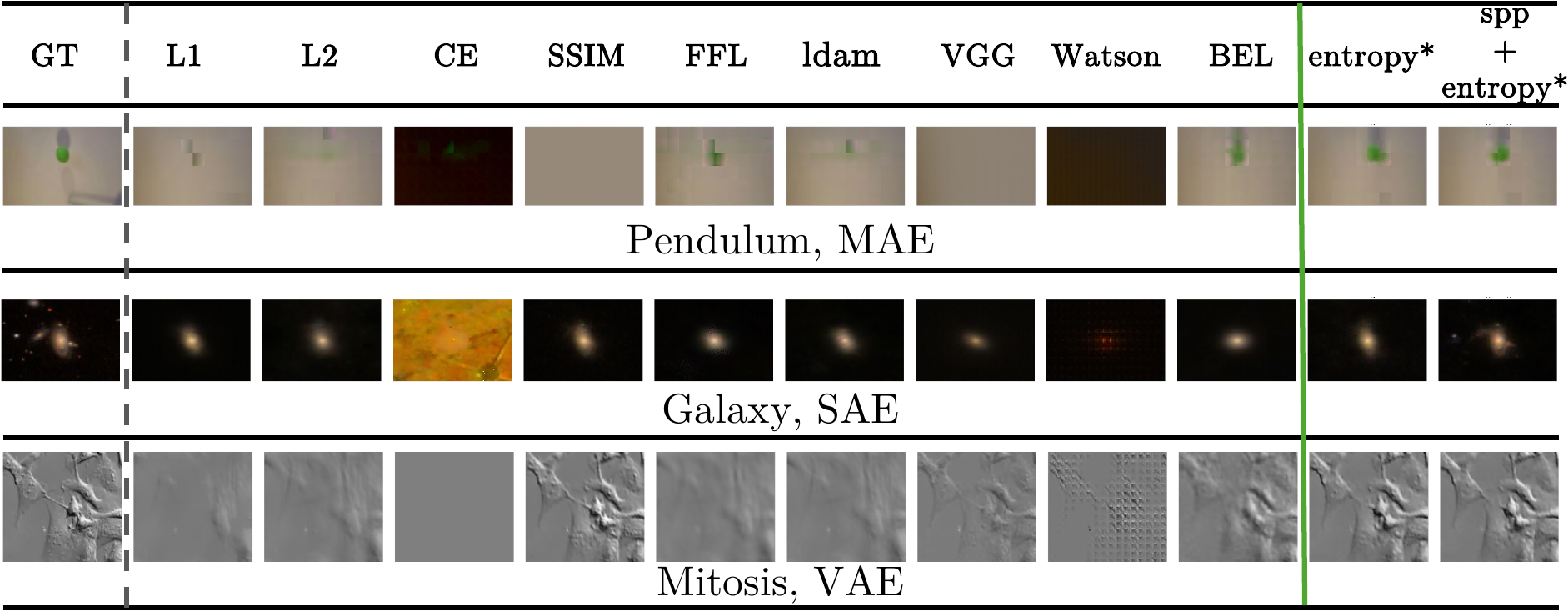}
    \caption{
    \textbf{Qualitative comparison under convergence: convergence toward the statistical mean vs.\ detail preservation.} For representative samples from \textit{Pendulum}, \textit{Galaxy}, and \textit{Mitosis} datasets, we show GT and reconstructions with baselines and our methods. The shown samples were \emph{randomly drawn} from the set; for clarity, we display the architecture per dataset where the effect is most legible, while other samples show similar tendencies. Pixel/perceptual baselines frequently converge to a dataset “mean” appearance—yielding blurred trajectories, deleting stars, and washed-out cellular boundaries, our losses retain more of the high-frequency structure and rare patterns, avoiding averaging and preserving relatively sharper edges and textures.
    }
    \vspace{-0.2cm}
    \label{fig:qualConver}
\end{figure}

Across three dataset containing repetetive dynamics (\textit{Pendulum}), structured textures (\textit{Galaxy}), and fine cellular morphology (\textit{Mitosis}), our entropy-based objectives  consistently deliver the best metrics (lowest MSE, highest PSNR) across all four architectures, while remaining competitive or superior in SSIM. On multiple experiments we see a consistent gain in PSNR while decreasing the MSE an order of magnitude (e.g. \textit{Pendulum-SAE}, \textit{Galaxy-AE}), indicating that maximizing information content prevents collapse to oversmoothed reconstructions. Some configurations present a better SSIM than ours, however our method is still comparable (e.g. ${<}0.025$ improvement on \textit{Pendulum-VAE}, \textit{Galaxy-SAE}, \textit{Mitosis-VAE}). Still our method retains a better performance in PSNR and MSE, suggesting sharper and perceptually faithful textures. 

On \textit{Mitosis-VAE}, entropy-based training preserves subtle cellular structure while substantially lowering distortion, outperforming baselines for all metrics. Our approach attains near-state-of-the-art SSIM with a clear better MSE/PSNR, highlighting a favorable position with respect to the competitors. Collectively, results imply that entropy maximization acts as a regularizer against convergence in repetitive or low-diversity regimes, improving generalization and detail preservation across architectures and data domains without task-specific tuning.

As illustrated in Fig.~\ref{fig:qualConver}, most baselines frequently converge toward a statistical mean appearance—producing blurred pendulum traces, smoothed galaxy arms, and softened cellular boundaries—whereas our entropy-based and spp$k$ objectives preserve high-frequency structure and rare patterns, yielding sharper trajectories, crisper spiral features, and better-defined mitotic contours that align with the quantitative gains reported above.

\paragraph{Sample propagation}

\begin{figure}[t]
    \centering
    \includegraphics[width=0.65\linewidth]{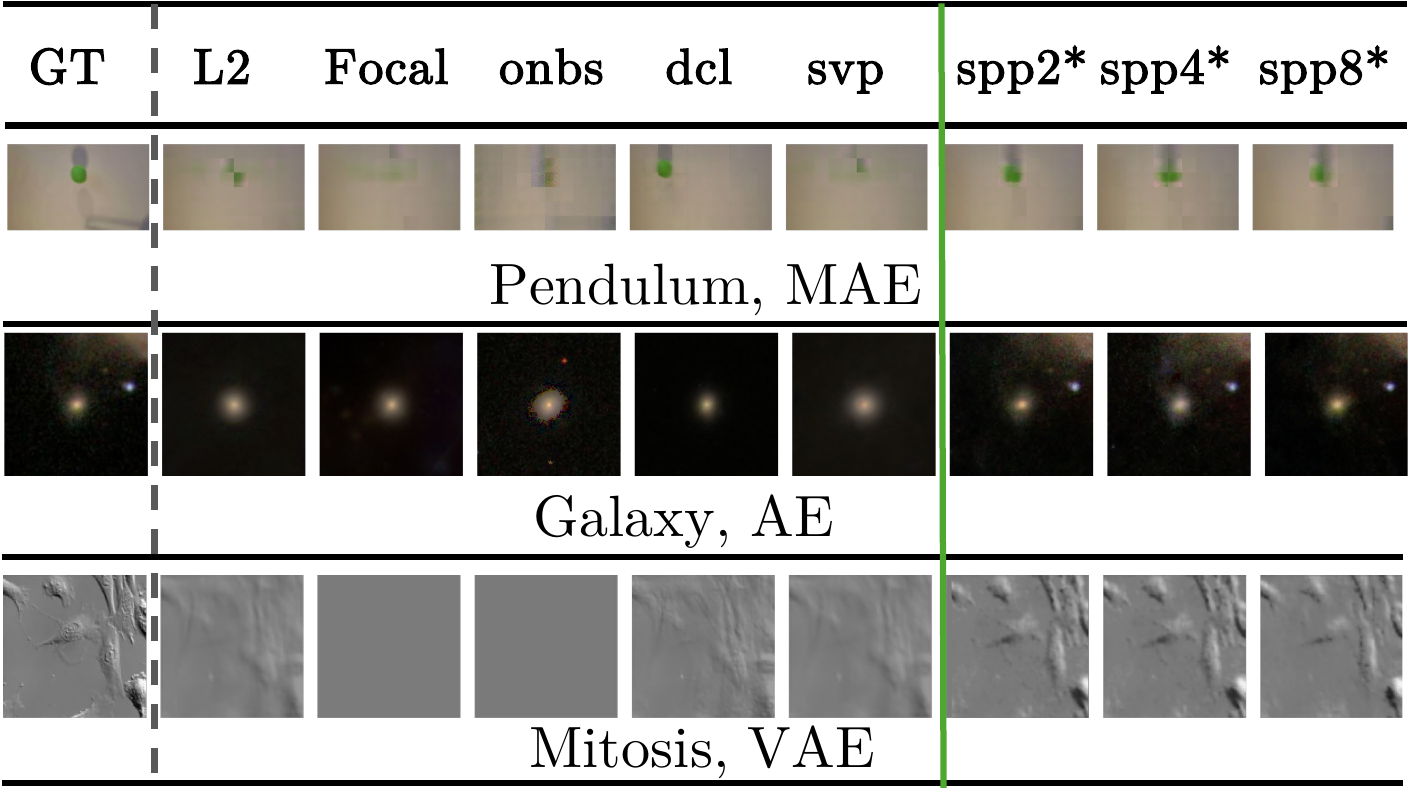}
    \caption{
    \textbf{Qualitative comparison with underrepresented-sample focusing (spp$k$).} Representative cases from the \textit{Pendulum}, \textit{Galaxy}, and \textit{Mitosis} datasets. We compare GT against L2/Focal/onbs/dcl/svp and our spp2*, spp4*, spp8* methods. The shown samples were \emph{randomly drawn} from the test sets; for clarity, we display the architecture per dataset where the effect is most legible, while other examples show similar tendencies. Pixel/perceptual baselines often smooth toward frequent patterns—fading spiral arms, indistinct mitotic contours, and blurred pendulum traces—whereas spp$k$ preserves rare/high-frequency structure, yielding relatively crisper edges and textures.
    }
    \label{fig:qualSpp}
    \vspace{-0.5cm}
\end{figure}

In this section, we evaluate the reconstruction performance of the AE models trained with our spp method. Spp is designed to improve learning from underrepresented samples, and we compare its results with baseline approaches. We show a qualitative analysis in Fig.~\ref{fig:qualSpp},  while the statistical results are summarized in Fig.~\ref{fig:summ_spp} the associated exhaustive tables can be found in the appendix~\ref{sec:spp_tables}.

Across all three datasets, spp$k$ improves average reconstruction. On repetitive dynamics (\textit{Pendulum}), larger memory leads to a better overall reconstruction, while on structured textures and fine details (\textit{Galaxy}, \textit{Mitosis}) smaller memories (spp4, spp8) dominate, suggesting that maintaining a buffer of rare patterns stabilizes learning and helps the model preserve high-frequency structure. 

Methodically, spp$k$ mitigates the bias towards frequent samples by increasing the effective sampling of underrepresented samples; this reduces blurring (better PSNR/MSE) and improves structural fidelity (SSIM). Overall, results indicate that targeting underrepresented samples is a principled way to more reliable outputs across data regimes and model families. As shown in Fig.~\ref{fig:qualSpp}, focusing on underrepresented samples also produces visibly sharper reconstructions and better retention of rare patterns.

\subsection{Downstream tasks}

Finally, in Table~\ref{tab:task_metrics}, we examine the benefits of applying our method across different tasks. First, we evaluate the image generation task used in BEL~\cite{bredell2023explicitly}. To ensure a fair comparison, we reimplemented their model and then built our method on top of it, we also report FID and LPIPS as standard generation metrics, and our approach outperforms the best referenced method. Second, we evaluate anomaly detection using the same dataset, setup, and metrics as the baseline in~\cite{collin2021improved}, achieving higher accuracy. Finally, we assess the impact of integrating our method into conventional AE training for physical parameter estimation. Using the same synthetic dataset as the Delfys baseline~\cite{garcia2024learning}, we addressed the inherent autoencoder limitations discussed in~\cite{garcia2024learning}, matching the performance of their best-reported models. Overall, results demonstrate that our framework consistently enhances performance across diverse domains in particular in tasks which suffer from spatial data imbalance.

\begin{table}
\centering
\caption{Downstream task performance (bold is better). In this table we compare three different task showing how our method can aid task by increasing the performance.}
\label{tab:task_metrics}
\scriptsize
\begin{tabular}{llcc}
\toprule
Task & Model & \multicolumn{2}{c}{Metrics} \\
\midrule
\multirow{3}{*}{Generation}
  &                 & FID & LPIPS \\
  & BEL~\cite{bredell2023explicitly}           & 280.5   & 0.3467 $\pm$ 0.0405    \\
  & \textbf{Ours}   & \textbf{221.8}  & \textbf{0.2826} $\pm$ \textbf{0.0512}     \\
\midrule
\multirow{3}{*}{\makecell{Anomaly\\detection}}
  &                 & Image AUROC & pixel AUROC \\
  & AESc~\cite{collin2021improved}            &  0.6017     & 0.7853         \\
  & \textbf{Ours}   & \textbf{0.6247}         & \textbf{0.7875}          \\
\midrule
\multirow{4}{*}{\makecell{Pendulum\\parameter\\estimation}}
  &                 & Abs. Err. (frequency) & Abs. Err. (damping) \\
  & Delfys~\cite{garcia2024learning}          & 0.0586 & 0.064 \\
  & AE              & 1.0016 & 3.32 \\
  & \textbf{Ours}            & \textbf{0.0516} & \textbf{0.002} \\
\bottomrule
\vspace{-0.3cm}
\end{tabular}\vspace{-1.0cm}
\end{table}
\section{Discussion}

Autoencoders are a core tool in unsupervised learning, enabling efficient representation learning across diverse tasks. However, the effect of spatial imbalance remains underexplored, despite its critical impact on downstream performance. Imbalanced spatial distributions can lead to suboptimal representations, reducing generalization and obscuring important variations. This study takes a first step toward addressing an often-overlooked challenge of data imbalance. By tackling this issue systematically, we show how unsupervised autoencoders can be improved without altering their underlying architectures.
 
In summary, our methods address the issue of spatial imbalance by improving the learning of underrepresented spatial locations, promoting a more balanced reconstruction. The self-entropy-based objective mitigates blurring and over-smoothing by maximizing information content, producing sharper, higher-fidelity reconstructions that preserve fine details across architectures. The sample propagation algorithm (spp) improves average accuracy and enhances retention of rare events while reducing focus on already learned features, leading to better performance in general across the whole dataset.

\textbf{Future Work:}
While our approach demonstrates significant improvements, a deeper understanding of data distribution behaviors is essential to determine when and how these techniques can be effectively applied on task-specific constraints.

\textbf{Limitations:}
Our study does not focus on a specific application domain, further validation in more complex and domain-specific setups is necessary to assess performance across diverse datasets and tasks.

\clearpage  


%
%
\bibliographystyle{splncs04}
\bibliography{main}
\clearpage
\appendix

\section{Technical Appendices and Supplementary Material}

\subsection{Mathematical details}

\subsubsection*{Probability calculation.}
\label{sec:apdx_pdf}
Pixel intensities are continuous on \([0,1]\). Thus, we define a histogram over a continuous variable, which estimates a \emph{density} (units: inverse intensity), not a discrete probability mass.
Normalizing counts by the batch size, \(c/B\), yields a \emph{frequency} for a bin of width \(\delta\); converting to a density requires dividing by \(\delta\) \vspace{-0.2cm}:
\[
\widehat{p}\ \approx\ \frac{c/B}{\delta}\ =\ \frac{J}{B}\,c.\vspace{-0.2cm}
\]
Dividing by \(B\) makes the estimate invariant to batch size; multiplying by \(J=1/\delta\) removes the dependence on the chosen bin width. Using a PMF-like \(c/B\) alone would shrink as bins get narrower (larger \(J\)), making \(-\log(\cdot)\) incomparable across choices of \(J\). The PDF-based surprise \(s=-\log\widehat{p}\) is therefore bin-width–invariant and consistently highlights statistically rare (high-surprise) pixels.

\subsection{Model Architectures}

Across all experiments, we relied on the architectures detailed in Table~\ref{tb:AE_net},~\ref{tb:VAE_net},~\ref{tb:SAE_net}, and ~\ref{tb:MAE_net}; which define the AE, VAE, SAE, and MAE networks respectively. While the overall structure remained consistent, the layer dimensionality and size were adapted to the specific dataset or method.


\begin{table}[]
\centering
  \setlength{\tabcolsep}{4pt}
  \renewcommand{\arraystretch}{0.8}  
  \caption{\textbf{MLP Autoencoder architecture.} 
  The input image of size $(c,h,w)$ is flattened to dimension $c{\cdot}h{\cdot}w$, where $c$ is the number of channels and $h,w$ are the spatial dimensions. 
  The encoder compresses the input into a latent vector of dimension $l$. 
  The reduction factors $d_i$ denote successive proportions used to shrink the layer dimensionality relative to the input size. The decoder mirrors this process to reconstruct.}
  \resizebox{0.6\columnwidth}{!}{%
  \begin{tabular}{llccc}
    \toprule
    \bf Stage & \bf Layer & \bf In & \bf Out & \bf Output Shape \\
    \midrule
    \multicolumn{5}{c}{\it Encoder} \\
    1 & Linear & $c\,h\,w$ & $c\,h\,w / d_1$ & $(\tfrac{c\,h\,w}{d_1})$ \\
      & ReLU   &           &                 & $(\tfrac{c\,h\,w}{d_1})$ \\
    2 & Linear & $c\,h\,w/d_1$ & $c\,h\,w / d_2$ & $(\tfrac{c\,h\,w}{d_2})$ \\
      & ReLU   &           &                 & $(\tfrac{c\,h\,w}{d_2})$ \\
    3 & Linear & $c\,h\,w/d_2$ & $l$ & $(l)$ \\
    \midrule
    \midrule
    \multicolumn{5}{c}{\it Decoder} \\
    1 & Linear & $l$ & $c\,h\,w / d_2$ & $(\tfrac{c\,h\,w}{d_2})$ \\
      & ReLU   &     &                  & $(\tfrac{c\,h\,w}{d_2})$ \\
    2 & Linear & $c\,h\,w/d_2$ & $c\,h\,w / d_1$ & $(\tfrac{c\,h\,w}{d_1})$ \\
      & ReLU   &     &                  & $(\tfrac{c\,h\,w}{d_1})$ \\
    3 & Linear & $c\,h\,w/d_1$ & $c\,h\,w$ & $(c,h,w)$ \\
      & Sigmoid &    &                  & $(c,h,w)$ \\
    \bottomrule
  \end{tabular}
  }
  \label{tb:AE_net}
\end{table}


\begin{table}[]
\centering
  \caption{\textbf{Variational Autoencoder (VAE) architecture.} 
  The input $(c,h,w)$ is encoded through three convolutional \emph{stages}, where each stage corresponds to a resolution-changing convolution (stride 2) followed by non-linearities. After flattening, fully connected layers reduce the representation to a latent distribution of dimension $l$, parameterized by mean ($\mu$) and variance ($\sigma$). Channel sizes $n_1,n_2,n_3$ are the number of feature maps at each stage, and $(h_i,w_i)$ are the spatial dimensions after stage $i$. The decoder mirrors this process with three transposed convolution \emph{stages}, reconstructing the image back to $(c,h,w)$. Intermediate fully connected layers are scaled by proportional factors $d_1$ and $d_2$ relative to $n_3$.}
\resizebox{0.85\textwidth}{!}{%
  \begin{tabular}{llcccl}
    \toprule
    \bf Stage & \bf Layer & \bf In & \bf Out & \bf Kernel / Stride / Pad & \bf Output Shape \\
    \midrule
    \multicolumn{6}{l}{\it Encoder} \\
    \multirow{2}{*}{1} & Conv2d   & $c$   & $n_1$ & $3\times3,\;s=2,\;p=1$ & $(n_1,h_1,w_1)$ \\
                        & ReLU     &       &       &                       & $(n_1,h_1,w_1)$ \\
    \multirow{2}{*}{2} & Conv2d   & $n_1$ & $n_2$ & $3\times3,\;s=2,\;p=1$ & $(n_2,h_2,w_2)$ \\
                        & ReLU     &       &       &                       & $(n_2,h_2,w_2)$ \\
    \multirow{2}{*}{3} & Conv2d   & $n_2$ & $n_3$ & $3\times3,\;s=2,\;p=1$ & $(n_3,h_3,w_3)$ \\
                        & ReLU     &       &       &                       & $(n_3,h_3,w_3)$ \\[6pt]
    \midrule
                        & Flatten  &       &       &                       & $(n_3 h_3 w_3)$ \\[3pt]
    \midrule
    \multicolumn{6}{l}{\textit{MLP Encoder}} \\
                        & Linear   & $n_3 h_3 w_3$ & $d_1 l$ & – & $(d_1 l)$ \\
                        & ReLU     &       &       &                       & $(d_1 l)$ \\
                        & Linear   & $d_1 l$ & $d_2 l$ & – & $(d_2 l)$ \\
                        & ReLU     &       &       &                       & $(d_2 l)$ \\
                        & Linear($\mu$)    & $d_2 l$ & $l$ & – & $(l)$ \\
                        & Linear($\sigma$) & $d_2 l$ & $l$ & – & $(l)$ \\
    \midrule
    \midrule
    \multicolumn{6}{l}{\it MLP Decoder} \\
                        & Linear   & $l$ & $d_2 l$ & – & $(d_2 l)$ \\
                        & ReLU     &     &           &   & $(d_2 l)$ \\
                        & Linear   & $d_2 l$ & $n_3 h_3 w_3$ & – & $(n_3 h_3 w_3)$ \\
                        & ReLU     &            &                  & $(n_3 h_3 w_3)$ \\[6pt]
    \midrule
                        & Reshape  &            &                  & – & $(n_3,h_3,w_3)$ \\[3pt]
    \midrule
    \multicolumn{6}{l}{\it Decoder} \\
    \multirow{2}{*}{1} & ConvTrans & $n_3$ & $n_2$ & $3\times3,\;s=2,\;p=1,\;op=1$ & $(n_2,h_2,w_2)$ \\
                        & ReLU      &       &       &                                & $(n_2,h_2,w_2)$ \\
    \multirow{2}{*}{2} & ConvTrans & $n_2$ & $n_1$ & $3\times3,\;s=2,\;p=1,\;op=1$ & $(n_1,h_1,w_1)$ \\
                        & ReLU      &       &       &                                & $(n_1,h_1,w_1)$ \\
    \multirow{2}{*}{3} & ConvTrans & $n_1$ & $c$   & $3\times3,\;s=2,\;p=1,\;op=1$ & $(c,h,w)$ \\
                        & ReLU      &       &       &                                & $(c,h,w)$ \\
                        & Conv2d    & $c$   & $c$   & $3\times3,\;s=1,\;p=1$         & $(c,h,w)$ \\
                        & ReLU      &       &       &                                & $(c,h,w)$ \\
                        & Conv2d    & $c$   & $c$   & $3\times3,\;s=1,\;p=1$         & $(c,h,w)$ \\
                        & Sigmoid   &       &       &                                & $(c,h,w)$ \\
    \bottomrule
  \end{tabular}
 }  
  \label{tb:VAE_net}
\end{table}

\begin{table}[]
\centering
\caption{\textbf{Sparse Autoencoder (SAE) architecture.} 
The encoder reduces the input image $(c,h,w)$ through three convolutional layers, producing feature maps $(n_3,h_0,w_0)$ with $h_0=h/8$ and $w_0=w/8$. This representation is flattened and mapped by an MLP into a latent vector of dimension $l$, with proportional factors $d_1$ and $d_2$ controlling the intermediate hidden sizes. A sparsity regularization is applied on $l$ to encourage compact codes. The decoder mirrors this process, expanding the latent code through an MLP, reshaping into convolutional feature maps, and progressively upsampling with transposed convolutions to reconstruct $(c,h,w)$.}
 \resizebox{0.80\textwidth}{!}{%
\begin{tabular}{lll}
\toprule
\textbf{Stage} & \textbf{Layer (Parameters)} & \textbf{Output Shape} \\
\midrule
\multicolumn{3}{l}{\textit{Encoder}} \\
Conv2d  & $(c \to n_1),\;3\times3,\;s=2,\;p=1$                  & $(n_1,\,h/2,\,w/2)$ \\
ReLU    & --                                                    & -- \\
Conv2d  & $(n_1 \to n_2),\;3\times3,\;s=2,\;p=1$                 & $(n_2,\,h/4,\,w/4)$ \\
ReLU    & --                                                    & -- \\
Conv2d  & $(n_2 \to n_3),\;3\times3,\;s=2,\;p=1$                 & $(n_3,\,h/8,\,w/8)$ \\
ReLU    & --                                                    & -- \\
\midrule
Flatten & --                                                    & $(n_3\,h_0\,w_0)$ \\
\midrule
\multicolumn{3}{l}{\textit{MLP Encoder}} \\
Linear  & $(n_3 h_0 w_0 \to d_1 l)$                             & $(d_1 l)$ \\
ReLU    & --                                                    & -- \\
Linear  & $(d_1 l \to d_2 l)$                                   & $(d_2 l)$ \\
ReLU    & --                                                    & -- \\
Linear  & $(d_2 l \to l)$                                       & $(l)$ \\
\midrule
\midrule
\multicolumn{3}{l}{\textit{MLP Decoder}} \\
Linear  & $(l \to d_2 l)$                                       & $(d_2 l)$ \\
ReLU    & --                                                    & -- \\
Linear  & $(d_2 l \to d_1 l)$                                   & $(d_1 l)$ \\
ReLU    & --                                                    & -- \\
Linear  & $(d_1 l \to n_3 h_0 w_0)$                             & $(n_3 h_0 w_0)$ \\
\midrule
\multicolumn{3}{l}{\textit{Decoder}} \\
ConvTranspose2d & $(n_3 \to n_2),\;3\times3,\;s=2,\;p=1,\;o_p=1$ & $(n_2,\,2h_0,\,2w_0)$ \\
ReLU            & --                                             & -- \\
ConvTranspose2d & $(n_2 \to n_1),\;3\times3,\;s=2,\;p=1,\;o_p=1$ & $(n_1,\,4h_0,\,4w_0)$ \\
ReLU            & --                                             & -- \\
ConvTranspose2d & $(n_1 \to c),\;3\times3,\;s=2,\;p=1,\;o_p=1$   & $(c,\,8h_0,\,8w_0)$ \\
ReLU            & --                                             & -- \\
Conv2d          & $(c \to c),\;3\times3,\;s=1,\;p=1$             & $(c,\,8h_0,\,8w_0)$ \\
ReLU            & --                                             & -- \\
Conv2d          & $(c \to c),\;3\times3,\;s=1,\;p=1$             & $(c,\,8h_0,\,8w_0)$ \\
Sigmoid         & --                                             & -- \\
\bottomrule
\end{tabular}
 }
\label{tb:SAE_net}
\end{table}

\begin{table}[]
  \centering
\caption{\textbf{Masked Autoencoder (MAE) architecture.} 
The model first embeds the input image $(N,C,H,W)$ into non‐overlapping patches of size $P\times P$, producing $L=(HW/P^2)$ tokens of embedding dimension $E$, where $C$ is the number of input channels. A learnable [CLS] token and fixed 2D sine–cosine positional embeddings are added. The encoder applies $d$ Transformer blocks with $h$ attention heads and MLP expansion ratio $r$, yielding latent representations of shape $(N,L+1,E)$. The decoder maps embeddings to dimension $D_{\text{dec}}$ (e.g., 16), adds mask tokens and positional encodings, and processes them with $d_{\text{dec}}$ Transformer blocks with $h_{\text{dec}}$ heads. Finally, a projection stage normalizes and linearly predicts patch pixels of size $P^2C$, which are reassembled into the reconstructed image $(N,C,H,W)$.}
\resizebox{0.85\textwidth}{!}{%
  \begin{tabular}{lllc}
    \toprule
    \textbf{Stage}         & \textbf{Module}            & \textbf{Configuration}                              & \textbf{Output Shape}              \\
    \midrule
    \multirow{2}{*}{Embed} & PatchEmbed                 & $P\!=\!16$, $D\!=\!E$ (e.g.\ 768/1024/1280), $C\!=\!3$ & $(N, L, E)$                        \\
                           & Add Pos.\ Embed (+CLS)     & fixed 2D sin‐cos                                    & $(N, L\!+\!1, E)$                  \\
    \midrule
    \multirow{2}{*}{Encoder}& Transformer Blocks ($\times d$) & heads $h$, MLP‐ratio $r$                         & $(N, L\!+\!1, E)$                  \\
                           & LayerNorm                  & $\mathrm{LayerNorm}(E)$                             & $(N, L\!+\!1, E)$                  \\
    \midrule
    \midrule
    \multirow{3}{*}{Decode} & Linear Embed               & $E \to D_{\text{dec}}$ (512)                        & $(N, L\!+\!1, D_{\text{dec}})$     \\
                           & Mask Token \& Pos.\ Embed  & fixed 2D sin‐cos                                    & $(N, L\!+\!1, D_{\text{dec}})$     \\
                           & Transformer Blocks ($\times d_{\text{dec}}$) & heads $h_{\text{dec}}$, MLP‐ratio $r$       & $(N, L\!+\!1, D_{\text{dec}})$     \\
    \midrule
    \multirow{3}{*}{Proj}   & LayerNorm                  & $\mathrm{LayerNorm}(D_{\text{dec}})$                & $(N, L\!+\!1, D_{\text{dec}})$     \\
                           & Linear Predict             & $D_{\text{dec}} \to P^2\!C$                         & $(N, L\!+\!1, P^2C)$               \\
                           & Remove CLS and Unpatchify  & —                                                   & $(N, C, H, W)$                     \\
    \bottomrule
  \end{tabular}
}
\label{tb:MAE_net}
\end{table}

\begin{table}[t]
\centering
\caption{Experimental configurations across conditions, datasets, and models.}
\resizebox{0.9\textwidth}{!}{%
\begin{tabular}{@{}lllll@{}}
\toprule
\textbf{Condition} & \textbf{Dataset} & \textbf{Model} & \textbf{Parameters} \\ \midrule
\multirow{12}{*}{Convergence} 
    & \multirow{4}{*}{Pendulum} 
        & AE  & $d_1 = 200,\; d_2 = 2000,\; l = 8$,\; $ \lambda = 1.0$ \\
    &   & VAE & $n_1 = 8,\; n_2 = 16,\; n_3 = 32,\; d_1 = 16,\; d_2 = 4,\; l = 2$,\; $\lambda = 1.0 $ \\
    &   & SAE & $n_1 = 8,\; n_2 = 16,\; n_3 = 32,\; d_1 = 16,\; d_2 = 4,\; l = 2$,\; $\lambda = 10.0 $ \\
    &   & MAE & patch size $= 16$, depth$=16$, heads $=16$, $\lambda = 10.0$ \\ \cmidrule(l){2-4}
    & \multirow{4}{*}{Mitosis} 
        & AE  & $d_1 = 200,\; d_2 = 2000,\; l = 8$,\; $\lambda = 1.0$  \\
    &   & VAE & $n_1 = 8,\; n_2 = 16,\; n_3 = 32,\; d_1 = 16,\; d_2 = 4,\; l = 2$,\; $\lambda = 1.0$  \\
    &   & SAE & $n_1 = 8,\; n_2 = 16,\; n_3 = 32,\; d_1 = 16,\; d_2 = 4,\; l = 2$,\; $\lambda = 20.0$  \\
    &   & MAE & patch size $= 16$, depth$=16$, heads$=16$, $\lambda = 10.0$ \\ \cmidrule(l){2-4}
    & \multirow{4}{*}{Galaxy} 
        & AE  & $d_1 = 200,\; d_2 = 2000,\; l = 8$ \\
    &   & VAE & $n_1 = 8,\; n_2 = 16,\; n_3 = 32,\; d_1 = 16,\; d_2 = 4,\; l = 2$ \\
    &   & SAE & $n_1 = 8,\; n_2 = 16,\; n_3 = 32,\; d_1 = 16,\; d_2 = 4,\; l = 2$ \\
    &   & MAE & patch size $= 16$, depth$=16$, heads $=16$, $\lambda = 10.0$ \\ \midrule
\multirow{12}{*}{Sample Unbalance} 
    & \multirow{4}{*}{Pendulum} 
        & AE  & $d_1 = 200,\; d_2 = 2000,\; l = 8$,\; $\lambda = 10.0$  \\
    &   & VAE & $n_1 = 8,\; n_2 = 16,\; n_3 = 32,\; d_1 = 16,\; d_2 = 4,\; l = 2$,\; $\lambda = 10.0$  \\
    &   & SAE & $n_1 = 8,\; n_2 = 16,\; n_3 = 32,\; d_1 = 16,\; d_2 = 4,\; l = 2$,\; $\lambda = 20.0 $ \\
    &   & MAE & patch size $= 16$, depth $=16$, heads $=16$, $\lambda = 10.0$ \\ \cmidrule(l){2-4}
    & \multirow{4}{*}{Mitosis} 
        & AE  & $d_1 = 200,\; d_2 = 2000,\; l = 8$,\; $\lambda = 1.0$  \\
    &   & VAE & $n_1 = 8,\; n_2 = 16,\; n_3 = 32,\; d_1 = 16,\; d_2 = 4,\; l = 2$,\; $\lambda = 1.0$  \\
    &   & SAE & $n_1 = 8,\; n_2 = 16,\; n_3 = 32,\; d_1 = 16,\; d_2 = 4,\; l = 2$,\; $\lambda = 1.0$  \\
    &   & MAE & patch size $= 16$, depth $=16$, heads $=16$, $\lambda = 10.0$ \\ \cmidrule(l){2-4}
    & \multirow{4}{*}{Galaxy} 
        & AE  & $d_1 = 200,\; d_2 = 2000,\; l = 8$,\; $\lambda = 1.0$  \\
    &   & VAE & $n_1 = 8,\; n_2 = 16,\; n_3 = 32,\; d_1 = 16,\; d_2 = 4,\; l = 2$,\; $\lambda = 1.0$  \\
    &   & SAE & $n_1 = 8,\; n_2 = 16,\; n_3 = 32,\; d_1 = 16,\; d_2 = 4,\; l = 2$,\; $\lambda = 1.0$  \\
    &   & MAE & patch size $= 16$, depth $=16$, heads $=16$, $\lambda = 10.0$ \\ 
\bottomrule
\end{tabular}}
\label{tb:parameterNet}
\end{table}

\subsubsection{Specification of Individual Experiments}
As discussed in the main paper, certain architectures have sufficient capacity to overfit the training set, thereby obscuring meaningful differences between reconstruction loss functions. To mitigate this effect and to more clearly expose the advantages of our approach, we systematically adjust dataset parameters in each experiment. The detailed configurations are reported here to ensure reproducibility and to provide a transparent account of how the evaluation setup emphasizes the strengths and limitations of different methods.  The parameter selection can be seen in table~\ref{tb:parameterNet}.

\subsection{Training details}

For the training we used a A40 GPU. The optimizer and learning rate was fixed for all of the experiment runs using Adam optimizer, learning rate $lr = 1.0e^{-3}$ and weight decay $wd = 1.0e^{-5}$.

\subsection{ Experimental results tables for real data}

On Tables~\ref{tb:PendulumResultsConvergence} to~\ref{tb:MitosisResultsConvergence} we show the quantitative results for the real-world data previously summarized on Fig.~\ref{fig:summ_conv} for the entropy experiments. Similarly, for the sample proagation experiments, the tables~\ref{tb:PendulumResultsSpp} to~\ref{tb:GalaxyResultsSpp} show the statistical results corresponding to Fig.~\ref{fig:summ_spp}.


\begin{table*}[t]
\centering
\caption{\textbf{Pendulum reconstruction: mitigating convergence on repetitive dynamics.} We compare loss functions across four autoencoder families (AE, VAE, SAE, MAE) on \textsc{Pendulum}. Rows marked with $*$ are our entropy–based objectives. Our losses consistently deliver the best distortion metrics (lowest MSE, highest PSNR) in every architecture: (i) \textbf{AE}: \emph{entropy} attains $1.66\times10^{-4}$ MSE / $38.45$,dB PSNR / $0.929$ SSIM, outperforming the strongest baseline (BEL) by $\approx$2$\times$ lower MSE, $+3.3$,dB PSNR, and $+0.040$ SSIM; (ii) \textbf{VAE}: \emph{entropy*} achieves the best MSE/PSNR ($1.02\times10^{-4}$, $40.00$,dB) with near–state–of–the–art SSIM (0.942), while the SSIM loss peaks in SSIM (0.964) at the cost of notably worse MSE/PSNR; (iii) \textbf{SAE}: \emph{spp+entropy*} yields the strongest MSE/PSNR ($1.12\times10^{-4}$, $39.69$,dB), and \emph{entropy*} attains the best SSIM ($0.948$); (iv) \textbf{MAE}: \emph{entropy*} is best on all three metrics ($5.93\times10^{-4}$, $32.90$,dB, $0.886$). The consistent gains—often multi–dB PSNR improvements and order–of–magnitude MSE reductions—indicate that maximizing information content during training prevents collapse on repetitive trajectories and preserves fine, non–repetitive details that pixel/perceptual objectives tend to smooth.}
\resizebox{0.7\textwidth}{!}{%
\begin{tabular}{clccc}\toprule
\multicolumn{1}{l}{\textbf{Model}} & \textbf{Loss} & \textbf{MSE} &  \textbf{PSNR} & \textbf{SSIM}  \\
\midrule
\multirow{{10}}{*}{AE}  & L1 & $1.126e^{-3} \pm 6.37e^{-4}$ & $3.078e^{1} \pm 4.06e^{0}$ & $8.777e^{-1} \pm 4.85e^{-2}$ \\
 & L2 & $8.224e^{-4} \pm 3.35e^{-4}$ & $3.127e^{1} \pm 2.01e^{0}$ & $8.548e^{-1} \pm 3.29e^{-2}$ \\
 & CE & $5.145e^{-3} \pm 4.95e^{-4}$ & $2.291e^{1} \pm 4.22e^{-1}$ & $7.467e^{-1} \pm 1.33e^{-2}$ \\
 & SSIM & $1.588e^{-3} \pm 1.29e^{-4}$ & $2.801e^{1} \pm 3.42e^{-1}$ & $8.507e^{-1} \pm 1.04e^{-2}$ \\
 & FFL & $8.188e^{-4} \pm 3.45e^{-4}$ & $3.128e^{1} \pm 1.98e^{0}$ & $8.583e^{-1} \pm 3.38e^{-2}$ \\
 & ldam & $6.611e^{-4} \pm 3.57e^{-4}$ & $3.229e^{1} \pm 2.00e^{0}$ & $8.693e^{-1} \pm 2.17e^{-2}$ \\ \cmidrule(l){2-5}
 & VGG & $1.596e^{-3} \pm 1.14e^{-4}$ & $2.798e^{1} \pm 3.03e^{-1}$ & $8.442e^{-1} \pm 1.04e^{-2}$ \\
 & Watson & $2.296e^{-3} \pm 1.91e^{-4}$ & $2.640e^{1} \pm 3.59e^{-1}$ & $6.934e^{-1} \pm 7.87e^{-3}$ \\
 & BEL & $3.359e^{-4} \pm 1.67e^{-4}$ & $3.511e^{1} \pm 1.69e^{0}$ & $8.890e^{-1} \pm 2.21e^{-2}$ \\ \cmidrule(l){2-5}
 & \textbf{entropy*} & $\bm{1.661e^{-4}} \pm 1.27e^{-4}$ & $\bm{3.845e^{1}} \pm 2.13e^{0}$ & $\bm{9.294e^{-1}} \pm 1.62e^{-2}$ \\
 & \textbf{spp+entropy*} & $9.833e^{-4} \pm 1.16e^{-4}$ & $3.010e^{1} \pm 5.08e^{-1}$ & $8.380e^{-1} \pm 1.87e^{-2}$ \\
 \midrule
 \multirow{{10}}{*}{VAE}  & L1 & $8.197e^{-4} \pm 9.51e^{-5}$ & $3.089e^{1} \pm 4.77e^{-1}$ & $8.846e^{-1} \pm 1.18e^{-2}$ \\
 & L2 & $1.499e^{-3} \pm 1.44e^{-4}$ & $2.826e^{1} \pm 4.13e^{-1}$ & $7.985e^{-1} \pm 1.12e^{-2}$ \\
 & CE & $6.391e^{-3} \pm 5.84e^{-4}$ & $2.196e^{1} \pm 4.02e^{-1}$ & $7.675e^{-1} \pm 1.19e^{-2}$ \\
 & SSIM & $2.283e^{-4} \pm 4.36e^{-5}$ & $3.648e^{1} \pm 7.01e^{-1}$ & $\bm{9.644e^{-1}} \pm 4.44e^{-3}$ \\
 & FFL & $1.663e^{-3} \pm 9.29e^{-5}$ & $2.780e^{1} \pm 2.40e^{-1}$ & $6.879e^{-1} \pm 5.93e^{-3}$ \\
 & ldam & $2.725e^{-3} \pm 1.81e^{-4}$ & $2.566e^{1} \pm 2.83e^{-1}$ & $8.145e^{-1} \pm 1.01e^{-2}$ \\ \cmidrule(l){2-5}
 & VGG & $4.871e^{-4} \pm 1.72e^{-4}$ & $3.329e^{1} \pm 1.07e^{0}$ & $9.318e^{-1} \pm 1.57e^{-2}$ \\
 & Watson & $2.685e^{-3} \pm 3.11e^{-4}$ & $2.574e^{1} \pm 4.95e^{-1}$ & $8.183e^{-1} \pm 1.26e^{-2}$ \\
 & BEL & $1.483e^{-3} \pm 1.32e^{-4}$ & $2.831e^{1} \pm 3.80e^{-1}$ & $8.102e^{-1} \pm 1.08e^{-2}$ \\ \cmidrule(l){2-5}
 & \textbf{entropy*} & $\bm{1.021e^{-4}} \pm 2.47e^{-5}$ & $\bm{4.000e^{1}} \pm 7.94e^{-1}$ & $9.419e^{-1} \pm 5.57e^{-3}$ \\
 & \textbf{spp+entropy*} & $1.064e^{-4} \pm 2.17e^{-5}$ & $3.979e^{1} \pm 6.72e^{-1}$ & $9.466e^{-1} \pm 5.72e^{-3}$ \\
\midrule
  \multirow{{10}}{*}{SAE}  & L1 & $1.625e^{-3} \pm 1.08e^{-4}$ & $2.790e^{1} \pm 2.97e^{-1}$ & $8.470e^{-1} \pm 8.77e^{-3}$ \\
 & L2 & $1.499e^{-3} \pm 1.36e^{-4}$ & $2.826e^{1} \pm 3.92e^{-1}$ & $8.029e^{-1} \pm 1.11e^{-2}$ \\
 & CE & $4.927e^{-3} \pm 4.10e^{-4}$ & $2.309e^{1} \pm 3.63e^{-1}$ & $7.851e^{-1} \pm 1.06e^{-2}$ \\
 & SSIM & $1.675e^{-3} \pm 1.19e^{-4}$ & $2.777e^{1} \pm 3.03e^{-1}$ & $8.439e^{-1} \pm 8.40e^{-3}$ \\
 & FFL & $1.659e^{-3} \pm 1.12e^{-4}$ & $2.781e^{1} \pm 2.92e^{-1}$ & $6.948e^{-1} \pm 6.48e^{-3}$ \\
 & ldam & $1.507e^{-3} \pm 1.46e^{-4}$ & $2.824e^{1} \pm 4.16e^{-1}$ & $7.932e^{-1} \pm 1.13e^{-2}$ \\ \cmidrule(l){2-5}
 & VGG & $1.639e^{-3} \pm 1.34e^{-4}$ & $2.787e^{1} \pm 3.40e^{-1}$ & $8.445e^{-1} \pm 9.75e^{-3}$ \\
 & Watson & $2.831e^{-3} \pm 1.72e^{-4}$ & $2.549e^{1} \pm 2.61e^{-1}$ & $8.347e^{-1} \pm 1.01e^{-2}$ \\
 & BEL & $1.649e^{-3} \pm 1.26e^{-4}$ & $2.784e^{1} \pm 3.27e^{-1}$ & $7.747e^{-1} \pm 1.04e^{-2}$ \\ \cmidrule(l){2-5}
 & \textbf{entropy*} & $1.620e^{-4} \pm 2.74e^{-5}$ & $3.795e^{1} \pm 5.93e^{-1}$ & $\bm{9.479e^{-1}} \pm 4.83e^{-3}$ \\
 & \textbf{spp+entropy*} & $\bm{1.124e^{-4}} \pm 5.10e^{-5}$ & $\bm{3.969e^{1}} \pm 1.11e^{0}$ & $9.460e^{-1} \pm 6.58e^{-3}$ \\ 
\midrule
\multirow{{10}}{*}{MAE}  & L1 & $1.423e^{-3} \pm 2.63e^{-4}$ & $2.856e^{1} \pm 9.13e^{-1}$ & $8.423e^{-1} \pm 2.22e^{-2}$ \\
 & L2 & $1.280e^{-3} \pm 2.41e^{-4}$ & $2.902e^{1} \pm 9.39e^{-1}$ & $7.989e^{-1} \pm 3.65e^{-2}$ \\
 & CE & $2.896e^{-1} \pm 4.87e^{-3}$ & $5.383e^{0} \pm 7.35e^{-2}$ & $1.673e^{-1} \pm 4.33e^{-2}$ \\
 & SSIM & $2.807e^{-3} \pm 1.88e^{-4}$ & $2.553e^{1} \pm 2.86e^{-1}$ & $8.231e^{-1} \pm 7.16e^{-3}$ \\
 & FFL & $1.392e^{-3} \pm 2.07e^{-4}$ & $2.862e^{1} \pm 7.19e^{-1}$ & $7.121e^{-1} \pm 6.71e^{-2}$ \\
 & ldam & $1.263e^{-3} \pm 2.52e^{-4}$ & $2.909e^{1} \pm 1.00e^{0}$ & $8.166e^{-1} \pm 3.30e^{-2}$ \\ \cmidrule(l){2-5}
 & VGG & $4.748e^{-3} \pm 3.79e^{-4}$ & $2.325e^{1} \pm 3.46e^{-1}$ & $7.622e^{-1} \pm 6.39e^{-3}$ \\
 & Watson & $1.730e^{-1} \pm 3.66e^{-3}$ & $7.620e^{0} \pm 9.24e^{-2}$ & $5.719e^{-1} \pm 4.70e^{-3}$ \\
 & BEL & $7.944e^{-4} \pm 3.79e^{-4}$ & $3.142e^{1} \pm 1.86e^{0}$ & $8.304e^{-1} \pm 3.44e^{-2}$ \\ \cmidrule(l){2-5}
 & \textbf{entropy*} & $\bm{ 5.928e^{-4} \pm 3.86e^{-4} }$ & $\bm{ 3.290e^{1} \pm 2.17e^{0} }$ & $\bm{ 8.856e^{-1} \pm 2.27e^{-2} }$ \\
 & \textbf{spp+entropy*} & $6.137e^{-4} \pm 3.89e^{-4}$ & $3.271e^{1} \pm 2.09e^{0}$ & $8.847e^{-1} \pm 1.79e^{-2}$ \\
\bottomrule
\end{tabular}
}
\label{tb:PendulumResultsConvergence}
\end{table*}
\begin{table*}[h]
\centering
\caption{
\textbf{Reconstruction performance on the \textsc{Galaxy} dataset.} We compare loss functions across four autoencoder families (AE, VAE, SAE, MAE). Rows marked with $*$ are our entropy–based objectives. Our method consistently achieves the best distortion metrics (lowest MSE, highest PSNR) in every architecture, while remaining competitive in SSIM (often within ${\sim}0.01$–$0.02$ of the SSIM-trained baseline). Concretely: (i) \textbf{AE}: \emph{entropy} attains the best MSE/PSNR ($9.57{\times}10^{-4}$, $31.59$,dB), improving upon the strongest non-ours baseline by ${\sim}38\%$ lower MSE and $+0.69$,dB PSNR; (ii) \textbf{VAE}: \emph{spp+entropy*} achieves the best MSE/PSNR ($8.78{\times}10^{-4}$, $31.79$,dB), a ${\sim}9\%$ MSE reduction and $+0.73$,dB over the best baseline; (iii) \textbf{SAE}: \emph{spp+entropy*} yields the best MSE/PSNR ($1.30{\times}10^{-3}$, $30.38$,dB), a ${\sim}32\%$ MSE drop and $+0.58$,dB; (iv) \textbf{MAE}: \emph{entropy*} provides the top PSNR ($30.01$,dB) with lower MSE ($1.41{\times}10^{-3}$), improving ${\sim}24\%$ in MSE and $+0.90$,dB. These gains indicate that maximizing information content mitigates convergence to oversmoothed solutions on structured astronomical textures, producing sharper, more faithful reconstructions across architectures.}
\resizebox{0.7\textwidth}{!}{%
\begin{tabular}{clccc}\toprule
\multicolumn{1}{l}{\textbf{Model}} & \textbf{Loss} & \textbf{MSE} &  \textbf{PSNR} & \textbf{SSIM}  \\
\midrule
\multirow{{10}}{*}{AE}  & L1 & $1.301e^{-3} \pm 1.61e^{-3}$ & $3.085e^{1} \pm 3.94e^{0}$ & $8.359e^{-1} \pm 4.64e^{-2}$ \\
 & L2 & $1.537e^{-3} \pm 1.45e^{-3}$ & $2.954e^{1} \pm 3.42e^{0}$ & $8.224e^{-1} \pm 4.92e^{-2}$ \\
 & CE & $1.707e^{-1} \pm 1.07e^{-2}$ & $7.686e^{0} \pm 2.70e^{-1}$ & $5.258e^{-1} \pm 3.48e^{-2}$ \\
 & SSIM & $1.268e^{-3} \pm 1.50e^{-3}$ & $3.090e^{1} \pm 3.91e^{0}$ & $\bm{8.527e^{-1}} \pm 4.53e^{-2}$ \\
 & FFL & $1.874e^{-3} \pm 2.21e^{-3}$ & $2.900e^{1} \pm 3.62e^{0}$ & $7.923e^{-1} \pm 6.07e^{-2}$ \\
 & ldam & $1.670e^{-3} \pm 1.82e^{-3}$ & $2.936e^{1} \pm 3.53e^{0}$ & $8.174e^{-1} \pm 5.03e^{-2}$ \\ \cmidrule(l){2-5}
 & VGG & $2.990e^{-3} \pm 5.33e^{-3}$ & $2.717e^{1} \pm 3.57e^{0}$ & $8.027e^{-1} \pm 4.76e^{-2}$ \\
 & Watson & $5.510e^{-3} \pm 9.70e^{-3}$ & $2.388e^{1} \pm 2.75e^{0}$ & $5.897e^{-1} \pm 8.58e^{-2}$ \\
 & BEL & $1.686e^{-3} \pm 1.89e^{-3}$ & $2.935e^{1} \pm 3.55e^{0}$ & $8.198e^{-1} \pm 5.00e^{-2}$ \\ \cmidrule(l){2-5}
 & \textbf{entropy*} & $\bm{9.567e^{-4}} \pm 9.13e^{-4}$ & $\bm{3.159e^{1}} \pm 3.38e^{0}$ & $8.377e^{-1} \pm 4.34e^{-2}$ \\
 & \textbf{spp+entropy*} & $1.048e^{-3} \pm 6.15e^{-4}$ & $3.026e^{1} \pm 1.96e^{0}$ & $8.137e^{-1} \pm 4.75e^{-2}$ \\
\midrule
\multirow{{10}}{*}{VAE}  & L1 & $1.268e^{-3} \pm 2.92e^{-3}$ & $3.045e^{1} \pm 3.12e^{0}$ & $8.123e^{-1} \pm 5.24e^{-2}$ \\
 & L2 & $9.686e^{-4} \pm 2.24e^{-3}$ & $3.106e^{1} \pm 2.15e^{0}$ & $7.954e^{-1} \pm 5.21e^{-2}$ \\
 & CE & $1.684e^{-1} \pm 1.31e^{-2}$ & $7.748e^{0} \pm 3.35e^{-1}$ & $6.202e^{-1} \pm 5.77e^{-2}$ \\
 & SSIM & $9.784e^{-4} \pm 2.85e^{-3}$ & $3.150e^{1} \pm 2.82e^{0}$ & $\bm{8.294e^{-1}} \pm 4.88e^{-2}$ \\
 & FFL & $1.283e^{-3} \pm 2.63e^{-3}$ & $2.990e^{1} \pm 2.42e^{0}$ & $7.656e^{-1} \pm 6.00e^{-2}$ \\
 & ldam & $1.004e^{-3} \pm 2.25e^{-3}$ & $3.090e^{1} \pm 2.18e^{0}$ & $7.891e^{-1} \pm 5.34e^{-2}$ \\ \cmidrule(l){2-5}
 & VGG & $2.412e^{-3} \pm 4.94e^{-3}$ & $2.796e^{1} \pm 3.24e^{0}$ & $8.022e^{-1} \pm 5.49e^{-2}$ \\
 & Watson & $1.886e^{-3} \pm 3.41e^{-3}$ & $2.876e^{1} \pm 3.20e^{0}$ & $6.720e^{-1} \pm 7.09e^{-2}$ \\
 & BEL & $3.127e^{-2} \pm 2.64e^{-3}$ & $1.506e^{1} \pm 3.43e^{-1}$ & $7.572e^{-1} \pm 6.00e^{-2}$ \\ \cmidrule(l){2-5}
 & \textbf{entropy*} & $8.951e^{-4} \pm 2.69e^{-3}$ & $3.163e^{1} \pm 2.47e^{0}$ & $8.196e^{-1} \pm 5.30e^{-2}$ \\
 & \textbf{spp+entropy*} & $\bm{8.780e^{-4}} \pm 2.71e^{-3}$ & $\bm{3.179e^{1}} \pm 2.42e^{0}$ & $8.171e^{-1} \pm 5.21e^{-2}$ \\
\midrule
\multirow{{10}}{*}{SAE}  & L1 & $1.913e^{-3} \pm 4.06e^{-3}$ & $2.952e^{1} \pm 3.88e^{0}$ & $8.069e^{-1} \pm 5.59e^{-2}$ \\
 & L2 & $2.224e^{-3} \pm 3.66e^{-3}$ & $2.834e^{1} \pm 3.48e^{0}$ & $7.757e^{-1} \pm 5.85e^{-2}$ \\
 & CE & $3.027e^{-1} \pm 1.55e^{-2}$ & $5.198e^{0} \pm 3.24e^{-1}$ & $6.933e^{-1} \pm 5.27e^{-2}$ \\
 & SSIM & $1.909e^{-3} \pm 4.45e^{-3}$ & $2.980e^{1} \pm 4.04e^{0}$ & $\bm{8.135e^{-1}} \pm 5.56e^{-2}$ \\
 & FFL & $2.320e^{-3} \pm 3.76e^{-3}$ & $2.807e^{1} \pm 3.36e^{0}$ & $7.583e^{-1} \pm 6.28e^{-2}$ \\
 & ldam & $2.228e^{-3} \pm 3.66e^{-3}$ & $2.833e^{1} \pm 3.48e^{0}$ & $7.795e^{-1} \pm 5.95e^{-2}$ \\ \cmidrule(l){2-5}
 & VGG & $3.534e^{-3} \pm 6.40e^{-3}$ & $2.620e^{1} \pm 3.34e^{0}$ & $7.730e^{-1} \pm 5.47e^{-2}$ \\
 & Watson & $ 1.476e^{-1} \pm 9.72e^{-3}$ & $ 8.324e^{0} \pm 4.09e^{-1}$ & $ 2.644e^{-1} \pm 1.04e^{-2}$ \\
 & BEL & $2.848e^{-3} \pm 6.37e^{-3}$ & $2.737e^{1} \pm 3.36e^{0}$ & $7.799e^{-1} \pm 5.22e^{-2}$ \\ \cmidrule(l){2-5}
 & \textbf{entropy*} & $1.374e^{-3} \pm 2.86e^{-3}$ & $3.025e^{1} \pm 3.34e^{0}$ & $8.103e^{-1} \pm 5.48e^{-2}$ \\
 & \textbf{spp+entropy*} & $\bm{1.302e^{-3}} \pm 2.89e^{-3}$ & $\bm{3.038e^{1}} \pm 3.20e^{0}$ & $8.097e^{-1} \pm 5.40e^{-2}$ \\ 
\midrule
 \multirow{{10}}{*}{MAE}  & L1 & $1.885e^{-3} \pm 3.40e^{-3}$ & $2.911e^{1} \pm 3.54e^{0}$ & $8.019e^{-1} \pm 5.63e^{-2}$ \\
 & L2 & $1.907e^{-3} \pm 3.09e^{-3}$ & $2.895e^{1} \pm 3.46e^{0}$ & $7.969e^{-1} \pm 5.55e^{-2}$ \\
 & CE & $3.720e^{-2} \pm 8.35e^{-3}$ & $1.438e^{1} \pm 8.25e^{-1}$ & $1.883e^{-1} \pm 2.15e^{-2}$ \\
 & SSIM & $1.924e^{-3} \pm 3.51e^{-3}$ & $2.903e^{1} \pm 3.54e^{0}$ & $\bm{ 8.053e^{-1} \pm 5.47e^{-2} }$ \\
 & FFL & $2.070e^{-3} \pm 3.25e^{-3}$ & $2.855e^{1} \pm 3.42e^{0}$ & $7.869e^{-1} \pm 5.63e^{-2}$ \\
 & ldam & $1.850e^{-3} \pm 2.55e^{-3}$ & $2.894e^{1} \pm 3.38e^{0}$ & $7.950e^{-1} \pm 5.60e^{-2}$ \\ \cmidrule(l){2-5}
 & VGG & $6.195e^{-3} \pm 7.06e^{-3}$ & $2.302e^{1} \pm 2.54e^{0}$ & $1.120e^{-1} \pm 4.62e^{-3}$ \\
 & Watson & $7.818e^{-3} \pm 7.99e^{-3}$ & $2.185e^{1} \pm 2.30e^{0}$ & $5.033e^{-2} \pm 5.19e^{-3}$ \\
 & BEL & $1.851e^{-3} \pm 2.61e^{-3}$ & $2.896e^{1} \pm 3.41e^{0}$ & $7.976e^{-1} \pm 5.54e^{-2}$ \\ \cmidrule(l){2-5}
 & \textbf{entropy*} & $1.409e^{-3} \pm 1.52e^{-3}$ & $\bm{ 3.001e^{1} \pm 3.41e^{0} }$ & $8.049e^{-1} \pm 5.38e^{-2}$ \\
 & \textbf{spp+entropy*} & $\bm{ 1.402e^{-3} \pm 1.51e^{-3} }$ & $2.998e^{1} \pm 3.35e^{0}$ & $8.053e^{-1} \pm 5.34e^{-2}$ \\
\bottomrule
\end{tabular}
}
\label{tb:GalaxyResultsConvergence}
\end{table*}
\begin{table*}[t]
\centering
\caption{\textbf{Reconstruction performance on the \textsc{Mitosis} dataset.} We compare loss functions across four autoencoder families (AE, VAE, SAE, MAE); rows marked with $*$ are our entropy–based objectives. \textbf{AE:} \emph{entropy} attains the best scores on all metrics ($6.65{\times}10^{-4}$ MSE, $32.10$,dB PSNR, $0.918$ SSIM), improving over the strongest non-ours baseline (ldam) by $\sim38\%$ lower MSE, $+2.17$,dB PSNR, and $+0.047$ SSIM. \textbf{VAE:} our \emph{spp+entropy*} achieves the best distortion ($1.05{\times}10^{-3}$ MSE, $29.95$,dB), edging the best baseline by $\sim 4.6\% $ MSE and $+0.16$,dB, while the SSIM-trained loss unsurprisingly peaks in SSIM ($0.909$). \textbf{SAE:} \emph{entropy*} is best on all three metrics ($3.01{\times}10^{-3}$, $25.49$,dB, $0.740$), yielding $\sim 36\%$ lower MSE, $+1.85$,dB PSNR, and $+0.044$ SSIM vs.\ the top baseline. \textbf{MAE:} \emph{spp+entropy*} leads across metrics ($3.88{\times}10^{-3}$, $24.30$,dB, $0.630$), a $\sim 29\% $ MSE drop and $+1.56$,dB PSNR over the best baseline with a $+0.020$ SSIM gain. Overall, maximizing information content mitigates oversmoothing and better preserves fine cellular structures, delivering consistently lower distortion and competitive–to–superior perceptual quality across architectures.}
\resizebox{0.7\textwidth}{!}{%
\begin{tabular}{clccc}\toprule
\multicolumn{1}{l}{\textbf{Model}} & \textbf{Loss} & \textbf{MSE} &  \textbf{PSNR} & \textbf{SSIM}  \\
\midrule
\multirow{{10}}{*}{AE}  & L1 & $1.546e^{-3} \pm 7.46e^{-4}$ & $2.848e^{1} \pm 1.72e^{0}$ & $8.588e^{-1} \pm 4.87e^{-2}$ \\
 & L2 & $1.366e^{-3} \pm 4.27e^{-4}$ & $2.883e^{1} \pm 1.25e^{0}$ & $8.449e^{-1} \pm 2.80e^{-2}$ \\
 & CE & $6.967e^{-3} \pm 1.25e^{-3}$ & $2.163e^{1} \pm 7.03e^{-1}$ & $8.927e^{-2} \pm 1.25e^{-2}$ \\
 & SSIM & $2.744e^{-3} \pm 1.43e^{-3}$ & $2.626e^{1} \pm 2.48e^{0}$ & $7.933e^{-1} \pm 8.16e^{-2}$ \\
 & FFL & $1.572e^{-3} \pm 5.27e^{-4}$ & $2.823e^{1} \pm 1.25e^{0}$ & $8.128e^{-1} \pm 3.71e^{-2}$ \\
 & ldam & $1.073e^{-3} \pm 4.05e^{-4}$ & $2.993e^{1} \pm 1.38e^{0}$ & $8.711e^{-1} \pm 2.81e^{-2}$ \\ \cmidrule(l){2-5}
 & VGG & $2.410e^{-3} \pm 1.19e^{-3}$ & $2.673e^{1} \pm 2.23e^{0}$ & $8.502e^{-1} \pm 5.56e^{-2}$ \\
 & Watson & $6.681e^{-3} \pm 1.39e^{-3}$ & $2.183e^{1} \pm 8.25e^{-1}$ & $4.361e^{-1} \pm 6.97e^{-2}$ \\
 & BEL & $2.422e^{-3} \pm 7.46e^{-4}$ & $2.636e^{1} \pm 1.33e^{0}$ & $7.628e^{-1} \pm 4.73e^{-2}$ \\ \cmidrule(l){2-5}
 & \textbf{entropy*} & $\bm{6.654e^{-4}} \pm 3.09e^{-4}$ & $\bm{3.210e^{1}} \pm 1.60e^{0}$ & $\bm{9.178e^{-1}} \pm 1.99e^{-2}$ \\
 & \textbf{spp+entropy*} & $1.478e^{-3} \pm 1.09e^{-4}$ & $2.831e^{1} \pm 2.84e^{-1}$ & $8.420e^{-1} \pm 2.65e^{-2}$ \\
\midrule
\multirow{{10}}{*}{VAE}  & L1 & $6.592e^{-3} \pm 1.63e^{-3}$ & $2.192e^{1} \pm 9.59e^{-1}$ & $5.223e^{-1} \pm 6.81e^{-2}$ \\
 & L2 & $6.319e^{-3} \pm 1.67e^{-3}$ & $2.212e^{1} \pm 1.03e^{0}$ & $4.992e^{-1} \pm 4.92e^{-2}$ \\
 & CE & $7.555e^{-3} \pm 1.43e^{-3}$ & $2.128e^{1} \pm 7.32e^{-1}$ & $5.300e^{-1} \pm 6.07e^{-2}$ \\
 & SSIM & $1.096e^{-3} \pm 3.37e^{-4}$ & $2.979e^{1} \pm 1.26e^{0}$ & $\bm{9.094e^{-1}} \pm 1.59e^{-2}$ \\
 & FFL & $6.396e^{-3} \pm 1.59e^{-3}$ & $2.206e^{1} \pm 9.67e^{-1}$ & $4.542e^{-1} \pm 4.10e^{-2}$ \\
 & ldam & $6.320e^{-3} \pm 1.61e^{-3}$ & $2.211e^{1} \pm 9.91e^{-1}$ & $4.898e^{-1} \pm 4.42e^{-2}$ \\ \cmidrule(l){2-5}
 & VGG & $3.774e^{-3} \pm 1.46e^{-3}$ & $2.456e^{1} \pm 1.69e^{0}$ & $7.807e^{-1} \pm 4.72e^{-2}$ \\
 & Watson & $8.002e^{-3} \pm 1.72e^{-3}$ & $2.105e^{1} \pm 8.30e^{-1}$ & $4.519e^{-1} \pm 7.21e^{-2}$ \\
 & BEL & $3.611e^{-3} \pm 1.15e^{-3}$ & $2.462e^{1} \pm 1.28e^{0}$ & $6.562e^{-1} \pm 5.03e^{-2}$ \\ \cmidrule(l){2-5}
 & \textbf{entropy*} & $1.054e^{-3} \pm 2.92e^{-4}$ & $2.992e^{1} \pm 1.11e^{0}$ & $8.798e^{-1} \pm 1.74e^{-2}$ \\
 & \textbf{spp+entropy*} & $\bm{1.046e^{-3}} \pm 2.92e^{-4}$ & $\bm{2.995e^{1}} \pm 1.12e^{0}$ & $8.736e^{-1} \pm 1.95e^{-2}$ \\
\midrule
 \multirow{{10}}{*}{SAE}  & L1 & $7.381e^{-3} \pm 1.43e^{-3}$ & $2.139e^{1} \pm 7.52e^{-1}$ & $5.305e^{-1} \pm 6.05e^{-2}$ \\
 & L2 & $6.312e^{-3} \pm 1.62e^{-3}$ & $2.212e^{1} \pm 9.97e^{-1}$ & $5.021e^{-1} \pm 4.73e^{-2}$ \\
 & CE & $7.345e^{-3} \pm 1.43e^{-3}$ & $2.141e^{1} \pm 7.53e^{-1}$ & $5.305e^{-1} \pm 6.05e^{-2}$ \\
 & SSIM & $4.709e^{-3} \pm 1.91e^{-3}$ & $2.364e^{1} \pm 1.82e^{0}$ & $6.965e^{-1} \pm 7.94e^{-2}$ \\
 & FFL & $6.385e^{-3} \pm 1.58e^{-3}$ & $2.206e^{1} \pm 9.63e^{-1}$ & $4.604e^{-1} \pm 4.16e^{-2}$ \\
 & ldam & $6.341e^{-3} \pm 1.58e^{-3}$ & $2.209e^{1} \pm 9.64e^{-1}$ & $4.682e^{-1} \pm 3.98e^{-2}$ \\ \cmidrule(l){2-5}
 & VGG & $7.207e^{-3} \pm 1.48e^{-3}$ & $2.150e^{1} \pm 7.92e^{-1}$ & $4.167e^{-1} \pm 4.68e^{-2}$ \\
 & Watson & $7.676e^{-3} \pm 1.42e^{-3}$ & $2.121e^{1} \pm 7.16e^{-1}$ & $3.781e^{-1} \pm 3.80e^{-2}$ \\
 & BEL & $6.295e^{-3} \pm 1.62e^{-3}$ & $2.213e^{1} \pm 1.00e^{0}$ & $5.029e^{-1} \pm 4.63e^{-2}$ \\ \cmidrule(l){2-5}
 & \textbf{entropy*} & $\bm{ 3.014e^{-3} \pm 1.12e^{-3} }$ & $\bm{ 2.549e^{1} \pm 1.54e^{0} }$ & $\bm{ 7.403e^{-1} \pm 4.58e^{-2} }$ \\
 & \textbf{spp+entropy*} & $3.925e^{-3} \pm 1.43e^{-3}$ & $2.435e^{1} \pm 1.59e^{0}$ & $6.840e^{-1} \pm 6.81e^{-2}$ \\ 
\midrule
 \multirow{{10}}{*}{MAE}  & L1 & $5.527e^{-3} \pm 1.21e^{-3}$ & $2.268e^{1} \pm 9.35e^{-1}$ & $5.553e^{-1} \pm 6.15e^{-2}$ \\
 & L2 & $5.546e^{-3} \pm 1.25e^{-3}$ & $2.266e^{1} \pm 9.22e^{-1}$ & $5.165e^{-1} \pm 5.30e^{-2}$ \\
 & CE & $2.433e^{-1} \pm 2.11e^{-3}$ & $6.139e^{0} \pm 3.74e^{-2}$ & $7.364e^{-2} \pm 5.77e^{-3}$ \\
 & SSIM & $5.492e^{-3} \pm 1.40e^{-3}$ & $2.274e^{1} \pm 1.08e^{0}$ & $6.095e^{-1} \pm 6.05e^{-2}$ \\
 & FFL & $5.721e^{-3} \pm 1.31e^{-3}$ & $2.253e^{1} \pm 9.30e^{-1}$ & $5.097e^{-1} \pm 5.43e^{-2}$ \\
 & ldam & $5.492e^{-3} \pm 1.25e^{-3}$ & $2.271e^{1} \pm 9.29e^{-1}$ & $5.319e^{-1} \pm 5.13e^{-2}$ \\ \cmidrule(l){2-5}
 & VGG & $7.683e^{-3} \pm 1.44e^{-3}$ & $2.121e^{1} \pm 7.30e^{-1}$ & $6.523e^{-2} \pm 4.06e^{-3}$ \\
 & Watson & $7.942e^{-3} \pm 1.53e^{-3}$ & $2.107e^{1} \pm 7.47e^{-1}$ & $3.009e^{-1} \pm 5.31e^{-2}$ \\
 & BEL & $5.272e^{-3} \pm 1.24e^{-3}$ & $2.290e^{1} \pm 1.02e^{0}$ & $5.222e^{-1} \pm 6.36e^{-2}$ \\ \cmidrule(l){2-5}
 & \textbf{entropy*} & $3.897e^{-3} \pm 1.07e^{-3}$ & $2.425e^{1} \pm 1.18e^{0}$ & $6.254e^{-1} \pm 5.43e^{-2}$ \\
 & \textbf{spp+entropy*} & $\bm{ 3.883e^{-3} \pm 1.15e^{-3} }$ & $\bm{ 2.430e^{1} \pm 1.27e^{0} }$ & $\bm{ 6.298e^{-1} \pm 6.06e^{-2} }$ \\
\bottomrule
\end{tabular}
}
\label{tb:MitosisResultsConvergence}
\end{table*}

\subsubsection*{}
\label{sec:spp_tables}

\begin{table*}[h]
\centering
\caption{
\textbf{Pendulum: effect of underrepresented-sample focusing on accuracy and robustness.} We compare losses across four architectures (AE, VAE, SAE, MAE). Rows with $*$ denote our \emph{spp$k$} objective that prioritizes underrepresented samples via a memory of size $k$. Considering both mean and variability (mean$\pm$std), \emph{spp} markedly improves average reconstruction on three architectures while keeping variance acceptable: \textbf{VAE}—\emph{spp2} attains the best distortion and perceptual quality (MSE$\downarrow$, PSNR$\uparrow$, SSIM$\uparrow$) with low–moderate spread; \textbf{SAE}—\emph{spp2*} achieves the strongest means across all metrics, with \emph{spp8*} slightly trading peak PSNR for reduced dispersion; \textbf{MAE}—\emph{spp2*} leads on all metrics and maintains tight standard deviations. For \textbf{AE}, focal loss offers the best means and small variance, while \emph{spp4*/spp8*} remain competitive in SSIM but do not surpass focal on averages. Across models, increasing memory (\emph{spp2}$\rightarrow$\emph{spp4}$\rightarrow$\emph{spp8}) generally smooths variance at a modest cost in mean performance, indicating a controllable mean–robustness trade-off. Overall, underrepresented-sample focusing improves average reconstruction quality where data imbalance is most impactful (VAE/SAE/MAE) and provides a tunable handle on stability via the standard deviation.
}
\resizebox{0.75\textwidth}{!}{%
\begin{tabular}{clccc}\toprule
\multicolumn{1}{l}{\textbf{Model}} & \textbf{Baseline} & \textbf{MSE} &  \textbf{PSNR} & \textbf{SSIM}  \\
\midrule
\multirow{{8}}{*}{AE}  & L2 & $1.292e^{-3} \pm 3.23e^{-4}$ & $2.905e^{1} \pm 1.27e^{0}$ & $8.129e^{-1} \pm 3.71e^{-2}$\\
 & Focal & $3.477e^{-4} \pm 1.56e^{-4}$ & $3.489e^{1} \pm 1.54e^{0}$ & $9.131e^{-1} \pm 1.61e^{-2}$\\
 & onbs & $2.533e^{-3} \pm 7.90e^{-4}$ & $2.636e^{1} \pm 2.22e^{0}$ & $7.679e^{-1} \pm 3.89e^{-2}$\\
 & dcl & $1.305e^{-3} \pm 3.71e^{-4}$ & $2.904e^{1} \pm 1.34e^{0}$ & $8.083e^{-1} \pm 3.63e^{-2}$\\
 & svp & $1.009e^{-3} \pm 3.70e^{-4}$ & $3.036e^{1} \pm 2.06e^{0}$ & $8.315e^{-1} \pm 4.12e^{-2}$\\ \cmidrule(l){2-5}
 & \textbf{spp2*} & $\bm{ 1.563e^{-3} \pm 1.49e^{-4} }$ & $2.808e^{1} \pm 4.26e^{-1}$ & $7.932e^{-1} \pm 1.77e^{-2}$\\
 & \textbf{spp4*} & $1.041e^{-3} \pm 5.05e^{-4}$ & $3.053e^{1} \pm 2.77e^{0}$ & $\bm{ 8.368e^{-1} \pm 4.85e^{-2} }$\\
 & \textbf{spp8*} & $1.147e^{-3} \pm 4.96e^{-4}$ & $\bm{ 3.012e^{1} \pm 2.94e^{0} }$ & $8.429e^{-1} \pm 4.67e^{-2}$\\
 \midrule
\multirow{{8}}{*}{VAE}  & L2 & $1.581e^{-3} \pm 9.09e^{-5}$ & $2.802e^{1} \pm 2.48e^{-1}$ & $7.832e^{-1} \pm 9.23e^{-3}$\\
 & Focal & $1.524e^{-3} \pm 9.91e^{-5}$ & $2.818e^{1} \pm 2.83e^{-1}$ & $7.953e^{-1} \pm 1.08e^{-2}$\\
 & onbs & $2.321e^{-3} \pm 3.76e^{-4}$ & $2.642e^{1} \pm 8.52e^{-1}$ & $\bm{ 6.651e^{-1} \pm 3.38e^{-2} }$\\
 & dcl & $1.685e^{-3} \pm 2.03e^{-4}$ & $2.776e^{1} \pm 5.08e^{-1}$ & $7.191e^{-1} \pm 2.20e^{-2}$\\
 & svp & $2.744e^{-3} \pm 1.83e^{-4}$ & $2.562e^{1} \pm 2.84e^{-1}$ & $8.143e^{-1} \pm 1.00e^{-2}$\\ \cmidrule(l){2-5}
 & \textbf{spp2*} & $\bm{ 1.806e^{-4} \pm 8.21e^{-5} }$ & $\bm{ 3.772e^{1} \pm 1.44e^{0} }$ & $9.310e^{-1} \pm 9.63e^{-3}$\\
 & \textbf{spp4*} & $2.754e^{-3} \pm 1.71e^{-4}$ & $2.561e^{1} \pm 2.67e^{-1}$ & $8.122e^{-1} \pm 1.02e^{-2}$\\
 & \textbf{spp8*} & $6.442e^{-4} \pm 1.15e^{-4}$ & $3.196e^{1} \pm 6.54e^{-1}$ & $8.583e^{-1} \pm 1.13e^{-2}$\\
\midrule
\multirow{{8}}{*}{SAE}  & L2 & $1.594e^{-3} \pm 9.58e^{-5}$ & $2.798e^{1} \pm 2.58e^{-1}$ & $7.757e^{-1} \pm 1.02e^{-2}$\\
 & Focal & $\bm{ 1.587e^{-3} \pm 8.24e^{-5} }$ & $2.800e^{1} \pm 2.24e^{-1}$ & $7.902e^{-1} \pm 8.33e^{-3}$\\
 & onbs & $2.540e^{-3} \pm 4.02e^{-4}$ & $2.602e^{1} \pm 8.46e^{-1}$ & $\bm{ 5.996e^{-1} \pm 2.71e^{-2} }$\\
 & dcl & $1.691e^{-3} \pm 1.68e^{-4}$ & $2.774e^{1} \pm 4.20e^{-1}$ & $7.413e^{-1} \pm 1.94e^{-2}$\\
 & svp & $1.529e^{-3} \pm 1.13e^{-4}$ & $2.817e^{1} \pm 3.18e^{-1}$ & $7.939e^{-1} \pm 1.17e^{-2}$\\ \cmidrule(l){2-5}
 & \textbf{spp2*} & $1.537e^{-4} \pm 1.79e^{-4}$ & $\bm{ 3.906e^{1} \pm 2.39e^{0} }$ & $9.420e^{-1} \pm 1.44e^{-2}$\\
 & \textbf{spp4*} & $7.563e^{-4} \pm 2.40e^{-4}$ & $3.138e^{1} \pm 1.11e^{0}$ & $8.656e^{-1} \pm 2.44e^{-2}$\\
 & \textbf{spp8*} & $3.080e^{-4} \pm 1.49e^{-4}$ & $3.545e^{1} \pm 1.57e^{0}$ & $9.253e^{-1} \pm 1.49e^{-2}$\\
\midrule
 \multirow{{8}}{*}{MAE}  & L2 & $1.271e^{-3} \pm 2.46e^{-4}$ & $2.906e^{1} \pm 9.62e^{-1}$ & $8.181e^{-1} \pm 3.08e^{-2}$\\
 & Focal & $1.283e^{-3} \pm 2.35e^{-4}$ & $2.900e^{1} \pm 9.07e^{-1}$ & $7.830e^{-1} \pm 3.85e^{-2}$\\
 & onbs & $2.102e^{-3} \pm 3.21e^{-4}$ & $2.683e^{1} \pm 7.05e^{-1}$ & $4.907e^{-1} \pm 1.38e^{-2}$\\
 & dcl & $1.331e^{-3} \pm 3.09e^{-4}$ & $2.889e^{1} \pm 1.12e^{0}$ & $7.887e^{-1} \pm 3.80e^{-2}$\\
 & svp & $1.294e^{-3} \pm 2.32e^{-4}$ & $2.896e^{1} \pm 8.92e^{-1}$ & $7.892e^{-1} \pm 3.82e^{-2}$\\ \cmidrule(l){2-5}
 & \textbf{spp2*} & $\bm{ 5.246e^{-4} \pm 3.28e^{-4} }$ & $\bm{ 3.335e^{1} \pm 2.00e^{0} }$ & $\bm{ 8.896e^{-1} \pm 1.53e^{-2} }$\\
 & \textbf{spp4*} & $6.862e^{-4} \pm 4.47e^{-4}$ & $3.227e^{1} \pm 2.18e^{0}$ & $8.837e^{-1} \pm 1.85e^{-2}$\\
 & \textbf{spp8*} & $5.593e^{-4} \pm 3.35e^{-4}$ & $3.299e^{1} \pm 1.81e^{0}$ & $8.823e^{-1} \pm 1.77e^{-2}$\\
\bottomrule
\end{tabular}
}
\label{tb:PendulumResultsSpp}
\end{table*}


\begin{table*}[h]
\centering
\caption{
\textbf{Mitosis: underrepresented-sample focusing improves averages with controllable variance.} We compare losses across four architectures (AE, VAE, SAE, MAE); rows with $*$ denote our \emph{spp$k$} objective (memory size $k$). Considering mean$\pm$std, \textbf{AE} benefits most from larger memory: \emph{spp8} attains the best MSE/PSNR/SSIM means, with a small increase in dispersion—indicating sharper reconstructions at slightly higher variability. \textbf{VAE} shows large gains with \emph{spp8*}/\emph{spp4*}, which substantially improve all three metrics while \emph{reducing} MSE variability versus pixel/perceptual baselines, suggesting more stable training on rare cell patterns. For \textbf{SAE}, \emph{spp4*} offers the strongest averages across metrics but with higher SSIM spread; \emph{spp8*} trades a bit of mean performance for tighter variance, exposing a tunable accuracy–robustness frontier. \textbf{MAE} mirrors this behavior: \emph{spp4*} yields the best means with standard deviations comparable to L2/Focal. Overall, prioritizing underrepresented samples consistently improves average reconstruction (MSE$\downarrow$, PSNR/SSIM$\uparrow$) and provides a controllable handle on robustness via the standard deviation, especially effective for VAE/MAE and with memory size governing the mean–variance trade-off.
}
\resizebox{0.75\textwidth}{!}{%
\begin{tabular}{clccc}\toprule
\multicolumn{1}{l}{\textbf{Model}} & \textbf{Baseline} & \textbf{MSE} &  \textbf{PSNR} & \textbf{SSIM}  \\
\midrule
\multirow{{8}}{*}{AE}  & L2 & $6.270e^{-3} \pm 1.54e^{-3}$ & $2.214e^{1} \pm 9.58e^{-1}$ & $4.983e^{-1} \pm 3.89e^{-2}$\\
 & Focal & $6.297e^{-3} \pm 1.52e^{-3}$ & $2.212e^{1} \pm 9.51e^{-1}$ & $4.966e^{-1} \pm 4.04e^{-2}$\\
 & onbs & $7.454e^{-3} \pm 1.58e^{-3}$ & $2.136e^{1} \pm 8.30e^{-1}$ & $9.822e^{-2} \pm 3.45e^{-2}$\\
 & dcl & $6.309e^{-3} \pm 1.46e^{-3}$ & $2.210e^{1} \pm 8.94e^{-1}$ & $4.901e^{-1} \pm 3.15e^{-2}$\\
 & svp & $6.303e^{-3} \pm 1.45e^{-3}$ & $2.210e^{1} \pm 8.91e^{-1}$ & $4.912e^{-1} \pm 3.28e^{-2}$\\ \cmidrule(l){2-5}
 & \textbf{spp2*} & $6.414e^{-3} \pm 1.56e^{-3}$ & $2.204e^{1} \pm 9.48e^{-1}$ & $5.136e^{-1} \pm 4.92e^{-2}$\\
 & \textbf{spp4*} & $6.420e^{-3} \pm 1.51e^{-3}$ & $2.203e^{1} \pm 9.19e^{-1}$ & $5.119e^{-1} \pm 4.49e^{-2}$\\
 & \textbf{spp8*} & $\bm{ 6.054e^{-3} \pm 1.59e^{-3} }$ & $\bm{ 2.232e^{1} \pm 1.11e^{0} }$ & $\bm{ 5.291e^{-1} \pm 5.71e^{-2} }$\\
\midrule
\multirow{{8}}{*}{ConvVAE}  & L2 & $6.328e^{-3} \pm 1.50e^{-3}$ & $2.209e^{1} \pm 9.34e^{-1}$ & $4.783e^{-1} \pm 3.57e^{-2}$\\
 & Focal & $6.363e^{-3} \pm 1.48e^{-3}$ & $2.207e^{1} \pm 9.15e^{-1}$ & $4.647e^{-1} \pm 3.43e^{-2}$\\
 & onbs & $7.342e^{-3} \pm 1.47e^{-3}$ & $2.142e^{1} \pm 7.75e^{-1}$ & $3.648e^{-1} \pm 5.29e^{-2}$\\
 & dcl & $6.350e^{-3} \pm 1.53e^{-3}$ & $2.208e^{1} \pm 9.46e^{-1}$ & $4.784e^{-1} \pm 3.76e^{-2}$\\
 & svp & $6.413e^{-3} \pm 1.45e^{-3}$ & $2.203e^{1} \pm 8.89e^{-1}$ & $4.458e^{-1} \pm 2.99e^{-2}$\\ \cmidrule(l){2-5}
 & \textbf{spp2*} & $2.085e^{-3} \pm 5.17e^{-4}$ & $2.693e^{1} \pm 9.98e^{-1}$ & $7.855e^{-1} \pm 2.38e^{-2}$\\
 & \textbf{spp4*} & $\bm{ 1.762e^{-3} \pm 5.11e^{-4} }$ & $\bm{ 2.771e^{1} \pm 1.20e^{0} }$ & $8.045e^{-1} \pm 2.79e^{-2}$\\
 & \textbf{spp8*} & $1.800e^{-3} \pm 4.83e^{-4}$ & $2.759e^{1} \pm 1.07e^{0}$ & $\bm{ 8.048e^{-1} \pm 2.28e^{-2} }$\\
\midrule
\multirow{{8}}{*}{SAE}  & L2 & $6.397e^{-3} \pm 1.35e^{-3}$ & $2.203e^{1} \pm 8.47e^{-1}$ & $4.535e^{-1} \pm 2.89e^{-2}$\\
 & Focal & $6.454e^{-3} \pm 1.30e^{-3}$ & $2.198e^{1} \pm 8.19e^{-1}$ & $4.302e^{-1} \pm 3.18e^{-2}$\\
 & onbs & $7.276e^{-3} \pm 1.44e^{-3}$ & $2.145e^{1} \pm 7.63e^{-1}$ & $3.984e^{-1} \pm 4.46e^{-2}$\\
 & dcl & $6.466e^{-3} \pm 1.40e^{-3}$ & $2.199e^{1} \pm 8.72e^{-1}$ & $4.381e^{-1} \pm 3.05e^{-2}$\\
 & svp & $6.432e^{-3} \pm 1.31e^{-3}$ & $2.200e^{1} \pm 8.18e^{-1}$ & $4.268e^{-1} \pm 2.62e^{-2}$\\ \cmidrule(l){2-5}
 & \textbf{spp2*} & $6.474e^{-3} \pm 1.35e^{-3}$ & $2.197e^{1} \pm 8.28e^{-1}$ & $4.901e^{-1} \pm 3.07e^{-2}$\\
 & \textbf{spp4*} & $\bm{ 4.715e^{-3} \pm 1.37e^{-3} }$ & $\bm{ 2.348e^{1} \pm 1.42e^{0} }$ & $\bm{ 6.343e^{-1} \pm 7.53e^{-2} }$\\
 & \textbf{spp8*} & $6.007e^{-3} \pm 1.05e^{-3}$ & $2.228e^{1} \pm 7.26e^{-1}$ & $5.137e^{-1} \pm 5.08e^{-2}$\\
\midrule
\multirow{{8}}{*}{MAE}  & L2 & $5.508e^{-3} \pm 1.25e^{-3}$ & $2.269e^{1} \pm 9.20e^{-1}$ & $5.370e^{-1} \pm 5.20e^{-2}$\\
 & Focal & $5.528e^{-3} \pm 1.23e^{-3}$ & $2.267e^{1} \pm 9.21e^{-1}$ & $5.325e^{-1} \pm 5.34e^{-2}$\\
 & onbs & $7.342e^{-3} \pm 1.63e^{-3}$ & $2.144e^{1} \pm 8.87e^{-1}$ & $3.954e^{-1} \pm 5.36e^{-2}$\\
 & dcl & $5.661e^{-3} \pm 1.37e^{-3}$ & $2.259e^{1} \pm 9.83e^{-1}$ & $5.087e^{-1} \pm 5.72e^{-2}$\\
 & svp & $5.616e^{-3} \pm 1.28e^{-3}$ & $2.261e^{1} \pm 9.26e^{-1}$ & $5.145e^{-1} \pm 5.64e^{-2}$\\ \cmidrule(l){2-5}
 & \textbf{spp2*} & $3.850e^{-3} \pm 1.05e^{-3}$ & $2.430e^{1} \pm 1.16e^{0}$ & $6.284e^{-1} \pm 5.22e^{-2}$\\
 & \textbf{spp4*} & $\bm{ 3.765e^{-3} \pm 1.09e^{-3} }$ & $\bm{ 2.442e^{1} \pm 1.23e^{0} }$ & $\bm{ 6.355e^{-1} \pm 5.62e^{-2} }$\\
 & \textbf{spp8*} & $3.799e^{-3} \pm 1.04e^{-3}$ & $2.436e^{1} \pm 1.16e^{0}$ & $6.327e^{-1} \pm 5.38e^{-2}$\\ 
\bottomrule
\end{tabular}
}
\label{tb:MitosisResultsSpp}
\end{table*}

\begin{table*}[h]
\centering
\caption{
\textbf{Galaxy: underrepresented-sample focusing improves accuracy and stabilizes training.} We compare losses across four architectures (AE, VAE, SAE, MAE); rows with $*$ denote our \emph{spp$k$} objective (memory size $k$). Considering mean$\pm$std, \textbf{AE} sees consistent gains with markedly lower variability: \emph{spp2} attains the best MSE/PSNR/SSIM among AEs ($9.19{\times}10^{-4}$, $31.71$,dB, $0.829$) with substantially smaller standard deviations than pixel/perceptual baselines. \textbf{VAE} benefits from larger memory: \emph{spp8*} yields the best MSE/PSNR ($8.13{\times}10^{-4}$, $31.91$,dB) and competitive SSIM (0.816), with variance comparable to or lower than L2/Focal. \textbf{SAE}: \emph{spp8*} provides the strongest triplet ($1.22{\times}10^{-3}$, $30.70$,dB, $0.818$) while reducing spread relative to baselines. \textbf{MAE}: \emph{spp8*} achieves the top PSNR/SSIM ($30.02$,dB, $0.805$), whereas \emph{spp4*} attains the lowest MSE ( $1.40{\times}10^{-3}$ ) with nearly identical variance—illustrating a controllable trade-off between mean accuracy and dispersion via memory size. Overall, prioritizing underrepresented samples improves average reconstruction (MSE$\downarrow$, PSNR/SSIM$\uparrow$) while keeping or lowering standard deviations, indicating better robustness across galaxy textures.
}
\resizebox{0.75\textwidth}{!}{%
\begin{tabular}{clccc}\toprule
\multicolumn{1}{l}{\textbf{Model}} & \textbf{Baseline} & \textbf{MSE} &  \textbf{PSNR} & \textbf{SSIM}  \\
\midrule
\multirow{{8}}{*}{AE}  & L2 & $1.465e^{-3} \pm 1.23e^{-4}$ & $2.836e^{1} \pm 3.62e^{-1}$ & $8.033e^{-1} \pm 1.38e^{-2}$\\
 & Focal & $1.449e^{-3} \pm 1.03e^{-4}$ & $2.840e^{1} \pm 3.06e^{-1}$ & $8.087e^{-1} \pm 1.11e^{-2}$\\
 & onbs & $2.476e^{-3} \pm 7.77e^{-4}$ & $2.645e^{1} \pm 2.16e^{0}$ & $7.642e^{-1} \pm 3.97e^{-2}$\\
 & dcl & $1.470e^{-3} \pm 1.50e^{-4}$ & $2.835e^{1} \pm 4.46e^{-1}$ & $7.841e^{-1} \pm 1.52e^{-2}$\\
 & svp & $1.460e^{-3} \pm 1.30e^{-4}$ & $2.837e^{1} \pm 3.85e^{-1}$ & $8.055e^{-1} \pm 1.42e^{-2}$\\ \cmidrule(l){2-5}
 & \textbf{spp2*} & $1.464e^{-3} \pm 3.01e^{-4}$ & $2.851e^{1} \pm 1.38e^{0}$ & $8.267e^{-1} \pm 2.24e^{-2}$\\
 & \textbf{spp4*} & $\bm{ 2.620e^{-4} \pm 2.81e^{-4} }$ & $\bm{ 3.706e^{1} \pm 2.90e^{0} }$ & $\bm{ 9.240e^{-1} \pm 3.00e^{-2} }$\\
 & \textbf{spp8*} & $6.298e^{-4} \pm 7.51e^{-4}$ & $3.455e^{1} \pm 4.52e^{0}$ & $9.110e^{-1} \pm 4.51e^{-2}$\\
\midrule
\multirow{{8}}{*}{VAE}  & L2 & $1.224e^{-3} \pm 2.12e^{-3}$ & $3.019e^{1} \pm 2.70e^{0}$ & $7.931e^{-1} \pm 5.30e^{-2}$\\
 & Focal & $1.468e^{-3} \pm 2.13e^{-3}$ & $2.928e^{1} \pm 2.58e^{0}$ & $7.664e^{-1} \pm 5.47e^{-2}$\\
 & onbs & $3.638e^{-3} \pm 7.04e^{-3}$ & $2.598e^{1} \pm 3.01e^{0}$ & $7.068e^{-1} \pm 4.80e^{-2}$\\
 & dcl & $2.238e^{-3} \pm 4.07e^{-3}$ & $2.851e^{1} \pm 3.61e^{0}$ & $7.886e^{-1} \pm 5.62e^{-2}$\\
 & svp & $1.165e^{-3} \pm 2.06e^{-3}$ & $3.025e^{1} \pm 2.41e^{0}$ & $7.840e^{-1} \pm 5.44e^{-2}$\\ \cmidrule(l){2-5}
 & \textbf{spp2*} & $9.785e^{-4} \pm 2.34e^{-3}$ & $3.116e^{1} \pm 2.57e^{0}$ & $8.114e^{-1} \pm 5.14e^{-2}$\\
 & \textbf{spp4*} & $8.682e^{-4} \pm 2.35e^{-3}$ & $3.166e^{1} \pm 2.44e^{0}$ & $\bm{ 8.163e^{-1} \pm 4.89e^{-2} }$\\
 & \textbf{spp8*} & $\bm{ 8.126e^{-4} \pm 1.86e^{-3} }$ & $\bm{ 3.191e^{1} \pm 2.45e^{0} }$ & $8.156e^{-1} \pm 4.93e^{-2}$\\
\midrule

\multirow{{8}}{*}{SAE}  & L2 & $2.179e^{-3} \pm 3.65e^{-3}$ & $2.853e^{1} \pm 3.58e^{0}$ & $7.945e^{-1} \pm 5.60e^{-2}$\\
 & Focal & $3.497e^{-3} \pm 5.38e^{-3}$ & $2.557e^{1} \pm 2.31e^{0}$ & $7.822e^{-1} \pm 5.13e^{-2}$\\
 & onbs & $3.353e^{-3} \pm 6.55e^{-3}$ & $2.637e^{1} \pm 3.03e^{0}$ & $7.151e^{-1} \pm 4.49e^{-2}$\\
 & dcl & $2.884e^{-3} \pm 6.40e^{-3}$ & $2.751e^{1} \pm 3.57e^{0}$ & $7.923e^{-1} \pm 5.26e^{-2}$\\
 & svp & $2.199e^{-3} \pm 3.72e^{-3}$ & $2.851e^{1} \pm 3.58e^{0}$ & $7.935e^{-1} \pm 5.54e^{-2}$\\ \cmidrule(l){2-5}
 & \textbf{spp2*} & $6.160e^{-3} \pm 6.95e^{-3}$ & $2.302e^{1} \pm 2.50e^{0}$ & $7.397e^{-1} \pm 5.73e^{-2}$\\
 & \textbf{spp4*} & $1.416e^{-3} \pm 2.71e^{-3}$ & $3.022e^{1} \pm 3.50e^{0}$ & $8.151e^{-1} \pm 5.22e^{-2}$\\
 & \textbf{spp8*} & $\bm{ 1.220e^{-3} \pm 2.65e^{-3} }$ & $\bm{ 3.070e^{1} \pm 3.27e^{0} }$ & $\bm{ 8.185e^{-1} \pm 5.11e^{-2} }$\\
\midrule
\multirow{{8}}{*}{MAE}  & L2 & $1.868e^{-3} \pm 2.62e^{-3}$ & $2.892e^{1} \pm 3.38e^{0}$ & $7.954e^{-1} \pm 5.64e^{-2}$\\
 & Focal & $2.401e^{-3} \pm 2.52e^{-3}$ & $2.727e^{1} \pm 2.72e^{0}$ & $7.503e^{-1} \pm 5.03e^{-2}$\\
 & onbs & $3.060e^{-3} \pm 6.32e^{-3}$ & $2.692e^{1} \pm 3.20e^{0}$ & $7.618e^{-1} \pm 5.05e^{-2}$\\
 & dcl & $2.123e^{-3} \pm 3.81e^{-3}$ & $2.864e^{1} \pm 3.55e^{0}$ & $7.887e^{-1} \pm 6.39e^{-2}$\\
 & svp & $1.903e^{-3} \pm 3.03e^{-3}$ & $2.894e^{1} \pm 3.43e^{0}$ & $7.907e^{-1} \pm 5.46e^{-2}$\\ \cmidrule(l){2-5}
 & \textbf{spp2*} & $1.408e^{-3} \pm 1.42e^{-3}$ & $2.994e^{1} \pm 3.35e^{0}$ & $8.050e^{-1} \pm 5.40e^{-2}$\\
 & \textbf{spp4*} & $\bm{ 1.399e^{-3} \pm 1.39e^{-3} }$ & $2.993e^{1} \pm 3.30e^{0}$ & $8.048e^{-1} \pm 5.33e^{-2}$\\
 & \textbf{spp8*} & $1.411e^{-3} \pm 1.55e^{-3}$ & $\bm{ 3.002e^{1} \pm 3.42e^{0} }$ & $\bm{ 8.052e^{-1} \pm 5.41e^{-2} }$\\
\bottomrule
\end{tabular}
}
\label{tb:GalaxyResultsSpp}
\end{table*}

\end{document}